\documentclass[10pt,twocolumn,letterpaper]{article}

\usepackage{wacv}              % To produce the CAMERA-READY version
\usepackage{graphicx}
\usepackage{amsmath}
\usepackage{amssymb}
\usepackage{booktabs}
\usepackage{multirow}

\usepackage[pagebackref,breaklinks,colorlinks]{hyperref}

\usepackage[capitalize]{cleveref}
\crefname{section}{Sec.}{Secs.}
\Crefname{section}{Section}{Sections}
\Crefname{table}{Table}{Tables}
\crefname{table}{Tab.}{Tabs.}

\def\wacvPaperID{531} % *** Enter the WACV Paper ID here
\def\confName{WACV}
\def\confYear{2024}

\begin{document}

\title{Single Frame Semantic Segmentation Using Multi-Modal Spherical Images}

\author{Suresh Guttikonda \qquad Jason Rambach \\
German Research Center for Artificial Intelligence (DFKI)  \\
{\tt\small \{suresh.guttikonda, jason.rambach\}@dfki.de}
}
\maketitle

%%%%%%%%% ABSTRACT
\begin{abstract}

In recent years, the research community has shown a lot of interest to panoramic images that offer a ${360}^{\circ}$ directional perspective. Multiple data modalities can be fed, and complimentary characteristics can be utilized for more robust and rich scene interpretation based on semantic segmentation, to fully realize the potential. Existing research, however, mostly concentrated on pinhole RGB-X semantic segmentation. In this study, we propose a transformer-based cross-modal fusion architecture to bridge the gap between multi-modal fusion and omnidirectional scene perception. We employ distortion-aware modules to address extreme object deformations and panorama distortions that result from equirectangular representation. Additionally, we conduct cross-modal interactions for feature rectification and information exchange before merging the features in order to communicate long-range contexts for bi-modal and tri-modal feature streams. In thorough tests using combinations of four different modality types in three indoor panoramic-view datasets, our technique achieved state-of-the-art mIoU performance: $60.60\%$ on Stanford2D3DS~\cite{DBLP:journals/corr/ArmeniSZS17} (RGB-HHA), $71.97\%$ Structured3D~\cite{DBLP:conf/eccv/ZhengZLTGZ20} (RGB-D-N), and $35.92\%$ Matterport3D~\cite{DBLP:conf/3dim/ChangDFHNSSZZ17} (RGB-D) \footnote{We plan to release all codes and trained models soon.}.

\end{abstract}

%%%%%%%%% BODY TEXT
\section{Introduction}
\label{sec:intro}

With the increased availability of affordable commercial 3D sensing devices, in recent years, researchers are more interested in working with omnidirectional images, also often referred to as ${360}^{\circ}$, panoramic, or spherical images. In contrast to pinhole cameras, the captured spherical images provide an ultra-wide ${360}^{\circ} \times {180}^{\circ}$ field-of-view (FoV) allowing for the capture of more detailed spatial information of the entire scene from a single frame~\cite{DBLP:conf/eccv/ZhangSTX14, DBLP:conf/icra/Guerrero-ViuFDG20}. Practical applications of such immersive and complete view perception include holistic and dense visual scene understanding~\cite{DBLP:journals/corr/abs-2205-10468}, augmented- and virtual reality (AR/VR)~\cite{DBLP:journals/pami/XuSWQHW19, DBLP:journals/tmm/QiaoXWB21}, autonomous driving~\cite{DBLP:conf/eccv/GaranderieAB18}, and robot navigation~\cite{DBLP:conf/cvpr/ChaplotS0020}.

Generally, spherical images are represented using equirectangular projection (ERP)~\cite{DBLP:conf/cvpr/0001ZRHS21} or cubemap projection (CP)~\cite{DBLP:conf/cvpr/WangYSCT20}, which introduces additional challenges like scene discontinuities, large image distortions, object deformations, and lack of open-source datasets with diverse real-world scenarios. While extensive research has been conducted on pinhole based learning methods~\cite{DBLP:conf/cvpr/LongSD15, DBLP:conf/iccv/CaoLLCTL21, DBLP:conf/nips/XieWYAAL21, DBLP:conf/cvpr/00010C00022, DBLP:journals/corr/abs-2203-04838}, approaches tailored for processing ultra-wide panoramic images and inherently accounting for spherical deformations remain ongoing research. Furthermore, the scarcity of labeled data, in indoor and outdoor scenarios, required for model training with panoramic images has slowed down the progress in this domain.

\begin{figure}
  \centering
  \includegraphics[width=\linewidth]{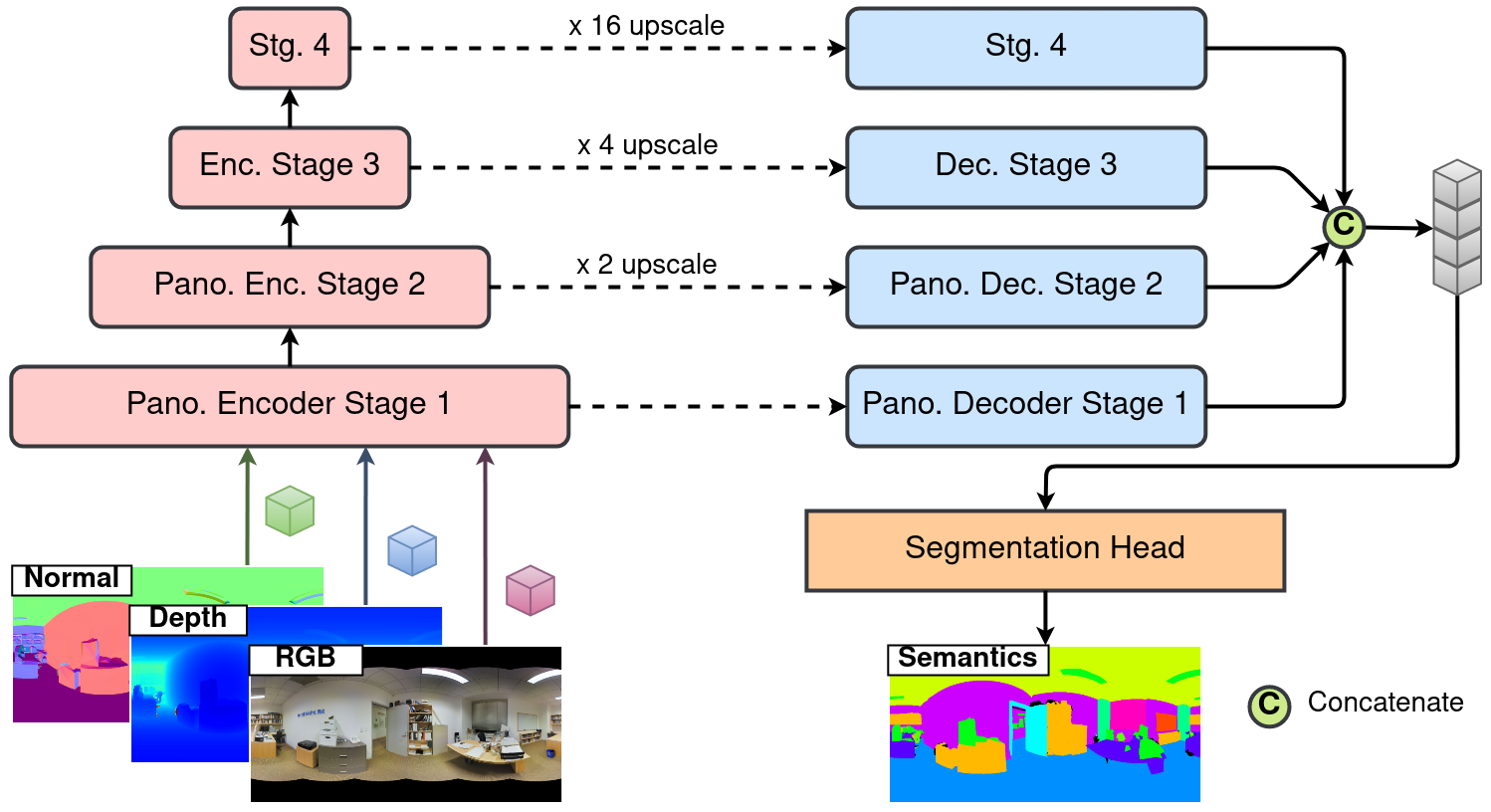}

  \caption{Overview of our multi-modal panoramic segmentation architecture. The inputs are an combination of \textbf{RGB}, \textbf{D}epth, and \textbf{N}ormals.}
  \label{fig:segmentationteaser}
\end{figure}

\begin{figure}
  \centering
  \begin{subfigure}{0.44\linewidth}
    % \fbox{\rule{0pt}{2in} \rule{.9\linewidth}{0pt}}
    \includegraphics[width=\linewidth]{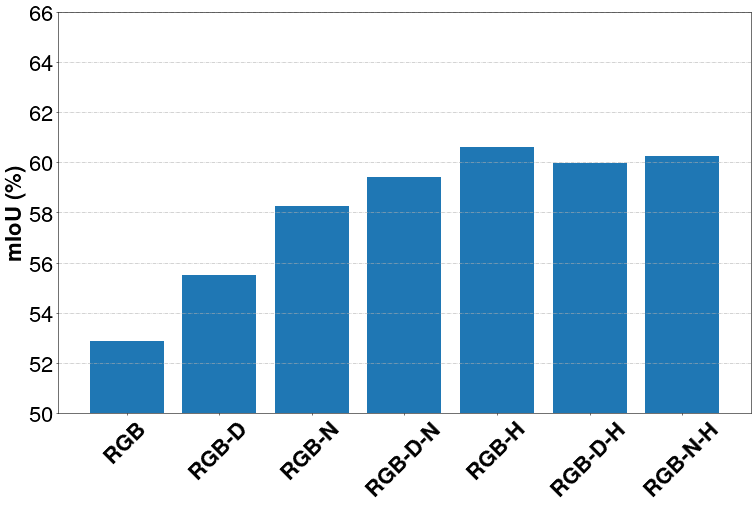}
    \caption{2D3DS~\cite{DBLP:journals/corr/ArmeniSZS17}}
  \end{subfigure}
  \begin{subfigure}{0.265\linewidth}
    % \fbox{\rule{0pt}{2in} \rule{.9\linewidth}{0pt}}
    \includegraphics[width=\linewidth]{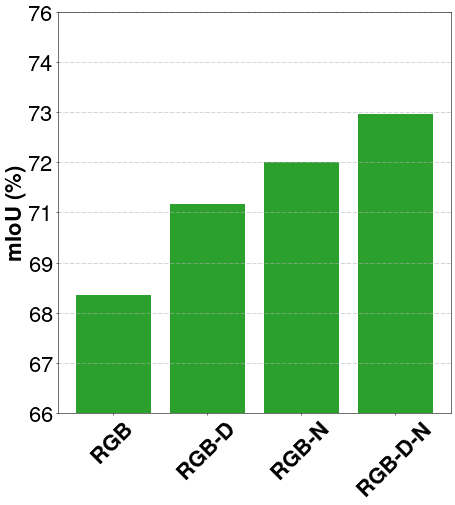}
    \caption{Struct3D~\cite{DBLP:conf/eccv/ZhengZLTGZ20}}
  \end{subfigure}
  \begin{subfigure}{0.265\linewidth}
    % \fbox{\rule{0pt}{2in} \rule{.9\linewidth}{0pt}}
    \includegraphics[width=\linewidth]{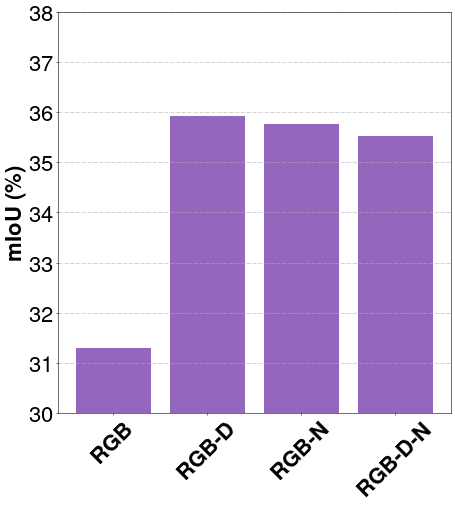}
    \caption{Mp3D~\cite{DBLP:conf/3dim/ChangDFHNSSZZ17}}
  \end{subfigure}
  \caption{Our cross-modal panoramic segmentation results with \textbf{RGB}, \textbf{D}epth, \textbf{N}ormals, and \textbf{H}HA combinations from Stanford2D3DS (\textit{left}), Structure3D (\textit{middle}) and Matterport3D (\textit{right}) datasets.}
  \label{fig:semgnetationresultsbarchart}
\end{figure}

While previous panorama segmentation techniques have attained state-of-the-art performance for \textbf{RGB}-only images, they do not take advantage of the complementary modalities to develop discriminative features in situations when it is difficult to discriminate only based on texture information. With comprehensive cross-modal interactions for \textbf{RGB}-\textbf{X} modality~\cite{DBLP:journals/corr/abs-2203-04838}, our work expands the current Trans4PASS+~\cite{DBLP:journals/corr/abs-2207-11860} methodology for multi-modal panoramic semantic segmentation. For the Stanford2D3DS~\cite{DBLP:journals/corr/ArmeniSZS17} dataset, we evaluate on $4$ distinct multi-modal semantic segmentation tasks, including \textbf{RGB}, \textbf{RGB}-\textbf{D}epth, \textbf{RGB}-\textbf{N}ormal, and \textbf{RGB}-\textbf{H}HA, and we reach a state-of-the-art $60.60\%$ with \textbf{RGB}-\textbf{H}HA semantic segmentation. We proposed a tri-modal fusion architecture and achieved top mIoU of $75.86\%$ on Structure3D~\cite{DBLP:conf/eccv/ZhengZLTGZ20} (RGB-D-N) and $39.26\%$ on Matterport3D~\cite{DBLP:conf/3dim/ChangDFHNSSZZ17} (RGB-D-N) for situations when HHA\footnote{\textbf{H}orizontal disparity, \textbf{H}eight above ground, and normal \textbf{A}ngle to the vertical axis~\cite{DBLP:conf/eccv/GuptaGAM14}} is not accessible. The performance of our system on the aforementioned indoor panoramic-view datasets is shown in~\cref{fig:semgnetationresultsbarchart}.

In summary, we provide the following contributions:
\begin{enumerate}
  \item We investigate multi-modal panoramic semantic segmentation in four types of sensory data combinations for the first time.
  \item We explore the multi-modal fusion paradigm in this study and introduce the tri-modal paradigm with cross-modal interactions for exploring texture, depth, and geometry information in panoramas.
  \item On three indoor panoramic datasets that include RGB, Depth, Normal, and HHA sensor data combinations, our technique provides state-of-the-art performance.
\end{enumerate}

%------------------------------------------------------------------------
\section{Related Work}
\label{sec:literacture}

\textbf{Semantic segmentation} An encoder-decoder paradigm with two stages is typically used in existing semantic segmentation designs~\cite{DBLP:journals/pami/BadrinarayananK17, DBLP:conf/eccv/ChenZPSA18}. A backbone \textit{encoder} module~\cite{DBLP:conf/cvpr/HeZRS16, DBLP:conf/cvpr/XieGDTH17, DBLP:conf/nips/GuoLHLC022} creates a series of feature maps in the earlier stage in order to capture high-level semantic data. Later, a \textit{decoder} module gradually extracts the spatial data from the feature maps. Recent research has focused on replacing convolutional backbones with transformer-based ones in light of the success of vision transformer (ViT) in imagine classification~\cite{DBLP:conf/iclr/DosovitskiyB0WZ21}. Early studies mostly concentrated on the Transformer encoder design~\cite{DBLP:conf/cvpr/ZhengLZZLWFFXT021, DBLP:conf/iccv/WangX0FSLL0021, DBLP:conf/iccv/LiuL00W0LG21, DBLP:conf/nips/ChuTWZRWXS21}, while later study avoided sophisticated decoders in favor of a lightweight All-MLP architecture~\cite{DBLP:conf/nips/XieWYAAL21}, which produced results with improved efficiency, accuracy, and robustness.

\textbf{Panoramic segmentation} Early methods for interpreting a picture holistically centered on using perspective image-based models in conjunction with distorted-mitigated wide-field of view images. A distortion-mitigated locally-planar image grid tangents to a subdivided icosahedron is Eder \etal~\cite{DBLP:conf/cvpr/EderSLF20} novel proposal for a tangent image spherical representation. Lee \etal~\cite{DBLP:conf/cvpr/LeeJYCY19}, on the other hand, uses a spherical polyhedron to symbolize comparable omni-directional perspectives. Recent studies~\cite{DBLP:journals/sivp/OrhanB22}, however, use distortion-aware modules in the network architecture to directly operate on equirectangular representation. Sun \etal~\cite{DBLP:conf/cvpr/0004SC21} suggests a discrete transformation for predicting dense features after an effective height compression module for latent feature representation. To improve the receptive field and learn the distortion distribution beforehand, Zheng \etal~\cite{DBLP:conf/wacv/ZhengLNLSZ23} combines the complimentary horizontal and vertical representation in the same line of research. In an encoder-decoder framework, Shen \etal~\cite{DBLP:conf/eccv/ShenLLNZZ22} introduces a brand-new panoramic transformer block to take the place of the conventional block. Modern panoramic distortion-aware and deformable modules~\cite{DBLP:conf/iccv/DaiQXLZHW17} have been added to the state-of-the-art UNet~\cite{DBLP:conf/miccai/RonnebergerFB15} and SegFormer~\cite{DBLP:conf/nips/XieWYAAL21} segmentation architectures to improve their performance in the spherical domain~\cite{DBLP:conf/icra/Guerrero-ViuFDG20, DBLP:journals/sivp/OrhanB22, DBLP:conf/cvpr/00010MRPS22, DBLP:journals/corr/abs-2207-11860}.

\textbf{Multimodal semantic segmentation} Fusion strategies leverage the advantages of several data sources and show notable performance improvements for image-based semantic segmentation~\cite{DBLP:conf/icip/HuYFW19, DBLP:journals/tip/ChenLWYC21}. The key contributions for comprehending \textbf{RGB}-\textbf{D} scenes concentrated on: 1) creating new layers or operators based on the geometric properties of \textbf{RGB}-\textbf{D} data~\cite{DBLP:conf/iccv/CaoLLCTL21, DBLP:conf/eccv/WangN18, DBLP:journals/tip/ChenLWYC21}, and 2) creating specialized architectures for combining the complimentary data streams in various stages~\cite{DBLP:conf/icip/HuYFW19, DBLP:conf/iccv/LeePH17, DBLP:conf/cvpr/0004SC21, DBLP:conf/eccv/ShenLLNZZ22}. When modalities other than depth maps are employed, these approaches perform less well because they were created exclusively for \textbf{RGB}-\textbf{D} modality~\cite{DBLP:conf/cvpr/0020ZLZHH21}. Recent studies have concentrated on establishing unique fusion algorithms for \textbf{RGB}-\textbf{X} semantic segmentation that are adaptable across various sensing modality combinations~\cite{DBLP:conf/cvpr/00010C00022, DBLP:journals/corr/abs-2203-04838, DBLP:journals/corr/abs-2303-01480}. In the omnidirectional realm, however, the integration of several modalities with cross-modal interactions is still an unresolved issue. The main issue in this scenario is to recognize the distorted and deformed geometric structures in the ultra-wide $360$-degree images while taking advantage of a variety of comprehensive complementing information. To jointly use the many sources of information from \textbf{RGB}, \textbf{D}epth, and \textbf{N}ormals equirectangular images, we propose our framework, which makes use of cross-modal interactions and panoramic perception abilities.

%------------------------------------------------------------------------
\section{Methodology}
\label{sec:framework}

~\Cref{sec:overview} provides a summary of the framework we propose for panoramic multi-modal semantic segmentation. Although our framework may be used for bi-modal and tri-modal input scenarios,for simplicity, we explain only the \textit{encoder} and \textit{decoder} architectures design for cross-modal (\textbf{RGB}-\textbf{D}epth-\textbf{N}ormals) panorama segmentation in ~\cref{sec:encoding} and ~\cref{sec:decoding}, respectively. Our design is based on Trans4PASS+~\cite{DBLP:journals/corr/abs-2207-11860} and uses an extension of CMX~\cite{DBLP:journals/corr/abs-2203-04838} for ternary modal streams feature extraction and fusion to learn object deformations and panoramic image distortions. We adopt a notation $\textbf{f}$ to represent multi-modal feature maps, \ie $\textbf{f} \in \{\textbf{f}_{rgb}, \textbf{f}_{depth}, \textbf{f}_{normal} \}$, in order to keep the notation simple and avoid the $l$ notation for inputs and outputs to network modules in the $l$-th encoder-decoder stage.

\begin{figure}
  \centering
  \includegraphics[width=\linewidth]{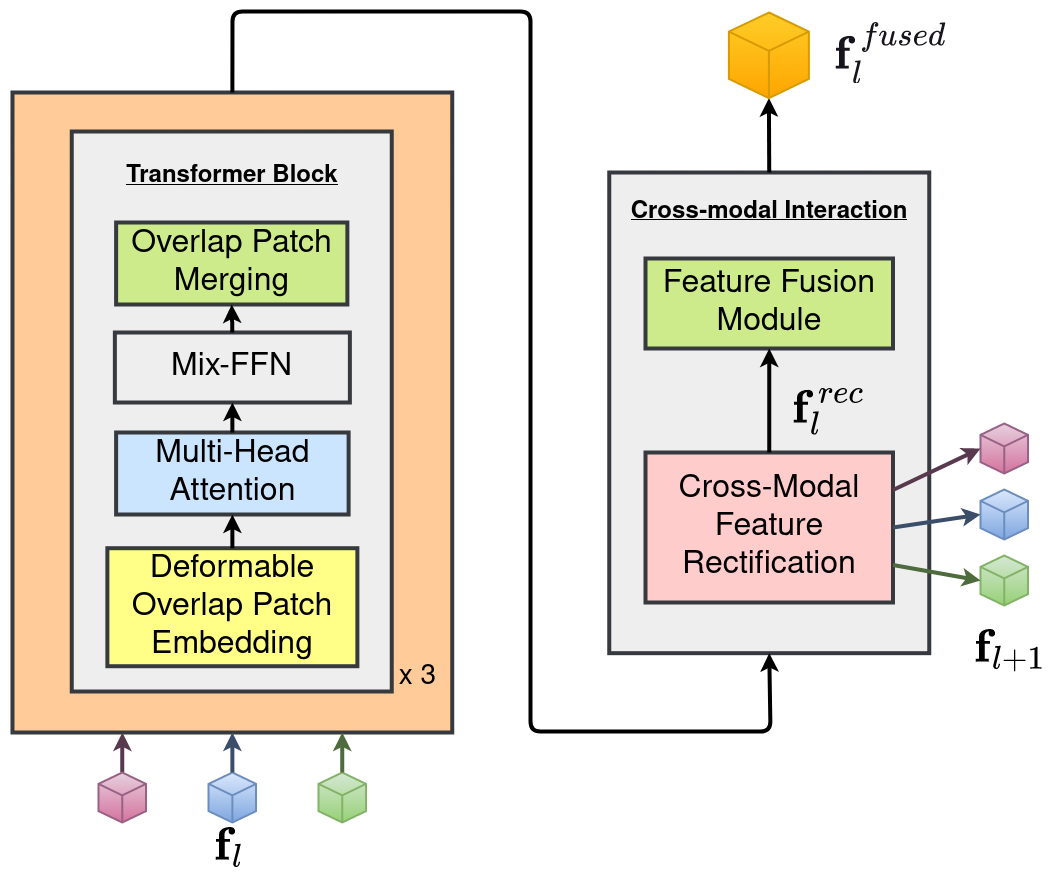}

  \caption{\textit{Panoramic encoder stage} to extract \textbf{RGB}, \textbf{D}epth, and \textbf{N}ormals features.}
  \label{fig:hierarchicalencoder}
\end{figure}

\subsection{Framework Overview}
\label{sec:overview}

In accordance with Xie \etal~\cite{DBLP:conf/nips/XieWYAAL21}, we proposed the multi-modal panoramic segmentation architecture depicted in~\cref{fig:segmentationteaser}. The $H \times W \times 3$ input image is first separated into patches. We provide panoramic hierarchical encoder stages to address the severe distortions in panoramas while allowing cross-modal interactions between \textbf{RGB}-\textbf{D}epth-\textbf{N}ormals patch features, as described in ~\cref{sec:encoding}. The encoder uses these patches as input to produce multi-level features at resolutions of $\{ 1/4, 1/8, 1/16, 1/32 \}$ of the original image. Finally, our panoramic decoder (refer~\cref{sec:decoding}) receives these multi-level features in order to predict the segmentation mask at a $ H \times W \times N_{class}$ resolution, where $N_{class}$ is the number of object categories.

\subsection{Panoramic Hierarchical Encoding}
\label{sec:encoding}

Each stage of our encoding process for extracting hierarchical characteristics is specifically designed and optimized for semantic segmentation.~\Cref{fig:hierarchicalencoder} illustrates how our architecture incorporates recently proposed Cross-modal Feature Rectification (CM-FRM) and Feature Fusion (FFM) modules~\cite{DBLP:journals/corr/abs-2203-04838} as well as Deformable Patch Embeddings (DPE) module~\cite{DBLP:conf/cvpr/00010MRPS22} to deal with the severe distortions in \textbf{RGB}, \textbf{D}epth, and \textbf{N}ormals panoramas caused by equirectangular representation.

\textbf{Deformable patch embedding} A typical Patch Embeddings (PE) module~\cite{DBLP:conf/iclr/DosovitskiyB0WZ21, DBLP:conf/nips/XieWYAAL21} divides an input image or feature map of size $\textbf{f} \in \mathbb{R}^{H \times W \times C_{in}}$ into a flattened 2D patch sequence of shape $s \times s$ each. In this patch, the position offset with respect to a location $ (i, j) $ is defined as $\boldsymbol{\Delta}_{( i, j )} \in \begin{bmatrix} \frac{-s}{2}, \frac{s}{2}\end{bmatrix} \times \begin{bmatrix} \frac{-s}{2}, \frac{s}{2} \end{bmatrix}$, where $ (i, j) \in [1, s] $. However, these fixed sample points fail to learn deformation-aware features and do not respect object shape distortions. To learn a data-dependent offset, we deploy a Deformable Patch Embeddings (DPE) module that was proposed by Zhang \etal~\cite{DBLP:conf/cvpr/00010MRPS22}. We formulate~\cref{eqn:dpe}, using the deformable convolution operation $g(.)$~\cite{DBLP:conf/iccv/DaiQXLZHW17} with a hyperparameter of $r = 4$.

\begin{equation}
    \boldsymbol{\Delta}^{DPE}_{( i, j )} = \begin{bmatrix} min(max(-\frac{H}{r}, g(\textbf{f})_{(i, j)}), \frac{H}{r}) \\
                                              min(max(-\frac{W}{r}, g(\textbf{f})_{(i, j)}), \frac{W}{r}) \end{bmatrix}
\label{eqn:dpe}
\end{equation}

\begin{figure}
  \centering
  \includegraphics[width=\linewidth]{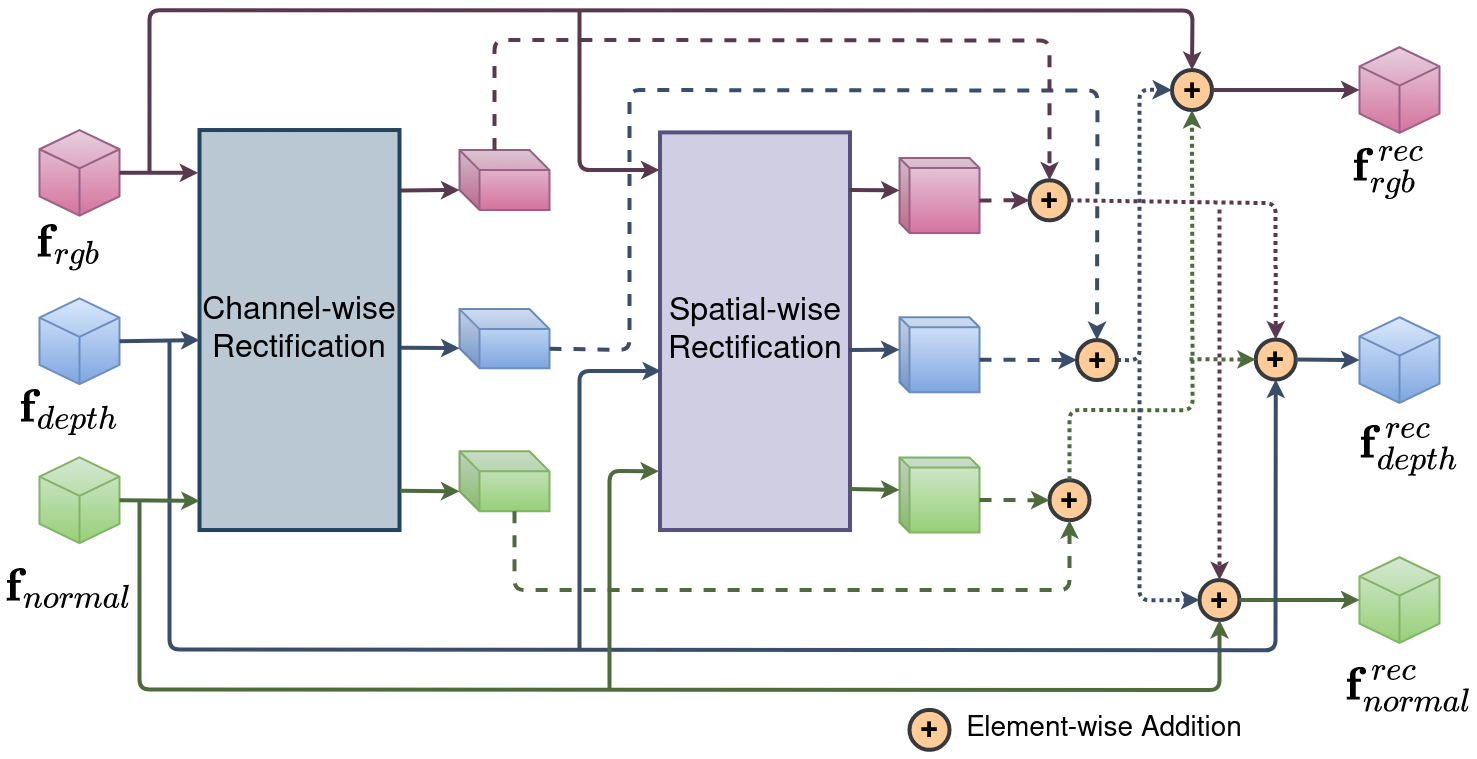}

  \caption{\textit{Cross-modal feature rectification module} to calibrate \textbf{RGB}, \textbf{D}epth, and \textbf{N}ormals features.}
  \label{fig:crossmodalfeaturerectification}
\end{figure}

\textbf{Cross-modal feature rectification} Measurements that are noisy are frequently present in the data from various complementing sensor modalities. By utilizing features from a different modality, the noisy information can be filtered and calibrated. Regarding this, Liu \etal~\cite{DBLP:journals/corr/abs-2203-04838} present a novel Cross-Modal Feature Rectification Module (CM-FRM) to execute feature rectification between parallel streams at each stage, throughout feature extraction process. In our work, we expand this calibration scheme using ternary features from \textbf{RGB}, \textbf{D}epth, and \textbf{N}ormals panorama stream, as seen in~\cref{fig:crossmodalfeaturerectification}. Our two-stage CM-FRM processes the input features channel- and spatial-wise to address noises and uncertainties in \textbf{RGB}-\textbf{D}epth-\textbf{N}ormals modalities, providing a comprehensive calibration for improved multi-modal feature extraction and interaction. While the spatial-wise rectification stage focuses on local calibration, the channel-wise rectification stage is more concerned with global calibrations. Hyperparameters $\lambda_{c}, \lambda_{s} = 0.5$ are utilized to rectify the noisy input multi-modal features as shown in~\cref{eqn:cmfrm} by using the channel $\textbf{f}^{~rec}_{channel}$ and spatial $\textbf{f}^{~rec}_{spatial}$ weights that have been obtained.

\begin{equation}
    \textbf{f}^{~rec} = \textbf{f} + \lambda_{c} \textbf{f}^{~rec}_{channel} + \lambda_{s} \textbf{f}^{~rec}_{spatial}
\label{eqn:cmfrm}
\end{equation}

\begin{figure}
  \centering
  \includegraphics[width=\linewidth]{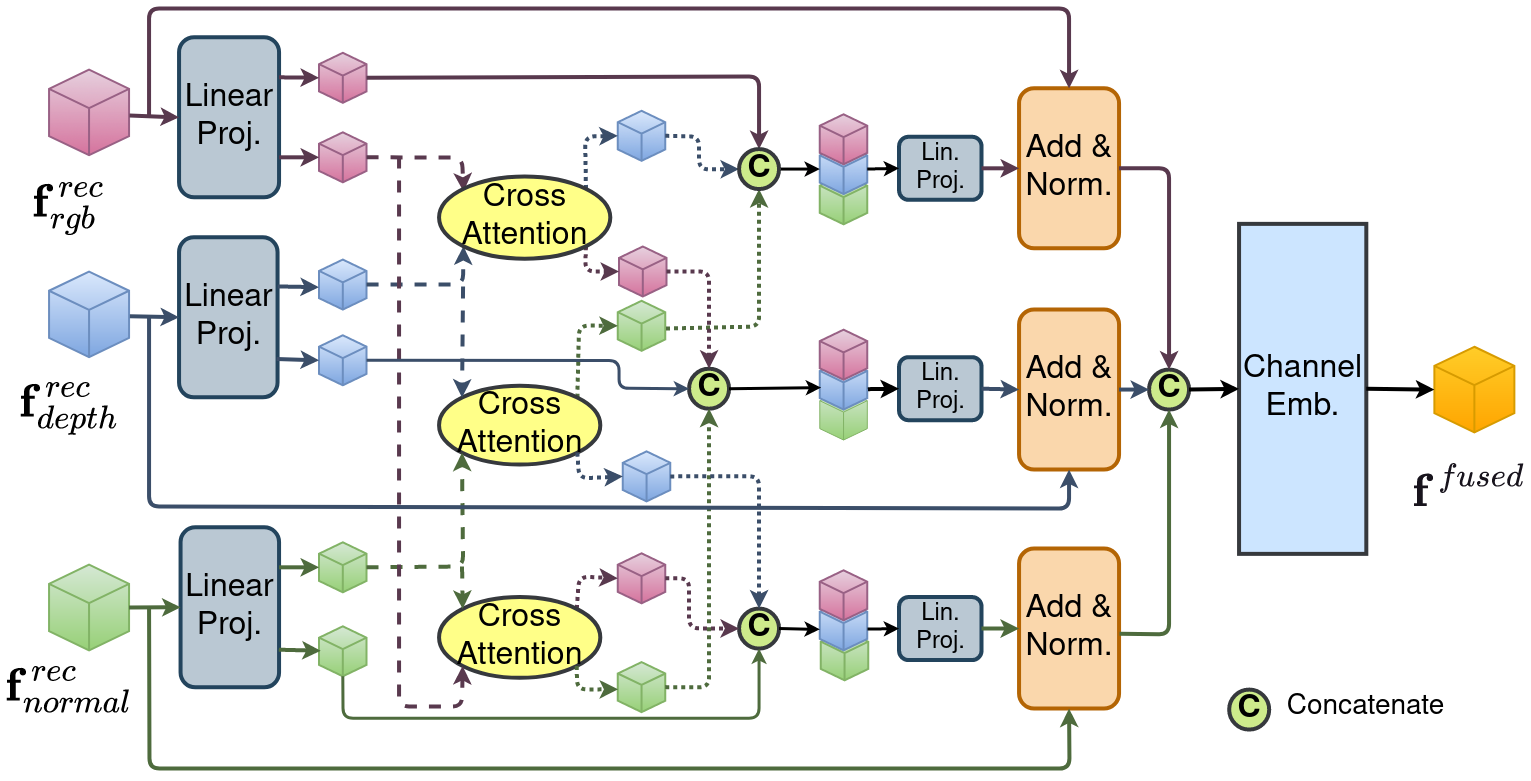}

  \caption{\textit{Cross-modal feature fusion module} to fuse \textbf{RGB}, \textbf{D}epth, and \textbf{N}ormals features.}
  \label{fig:crossmodalfeaturefusion}
\end{figure}

\textbf{Cross-modal feature fusion} To improve information interaction and combine the features into a single feature map the rectified multi-modal feature maps $\textbf{f}^{~rec}$ are passed through a two-stage Feature Fusion Module (FFM) at the end of each encoder stage. As seen in~\cref{fig:crossmodalfeaturefusion}, we use a ternary multi-head cross-attention mechanism to expand Liu \etal~\cite{DBLP:journals/corr/abs-2203-04838} information sharing stage by allowing for global information flow between the \textbf{RGB}, \textbf{D}epth, and \textbf{N}ormals modalities. In the fusion stage, a channel embedding~\cite{DBLP:journals/corr/abs-2203-04838} is utilized to combine ternary features to $\textbf{f}^{~fused}$ and passed through the decoding step for semantics prediction.

\subsection{Panoramic Token Mixer Decoder}
\label{sec:decoding}

\begin{figure}
  \centering
  \includegraphics[width=\linewidth]{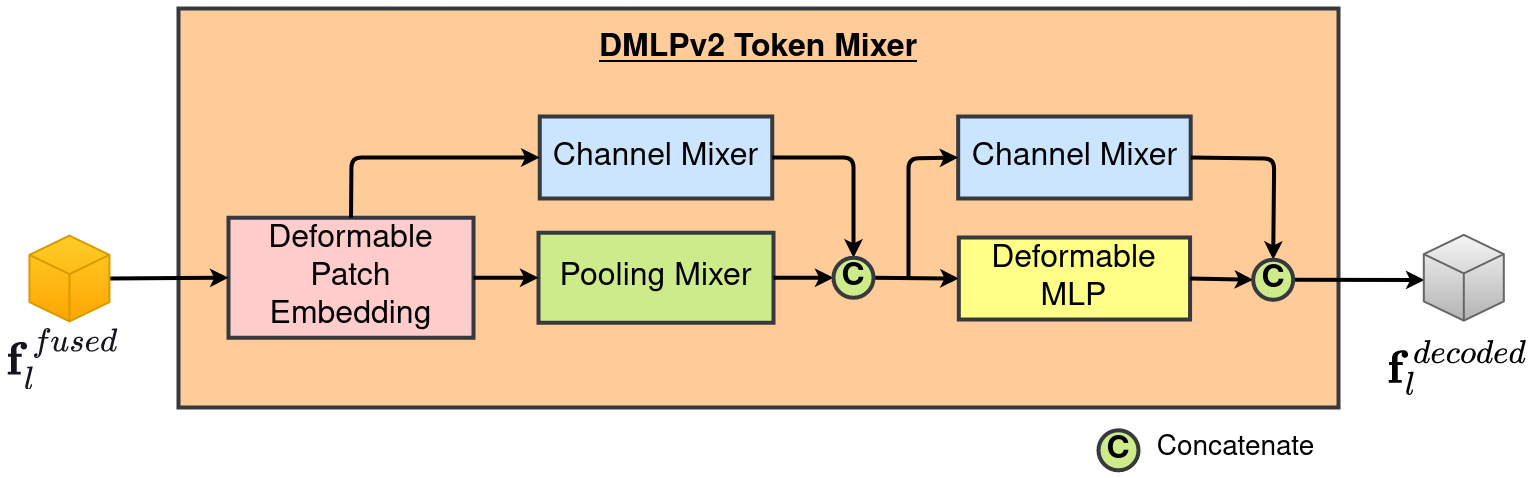}

  \caption{\textit{Panoramic decoder stage} with fused features from \textbf{RGB}, \textbf{D}epth, and \textbf{N}ormals modalities.}
  \label{fig:dmlpv2tokenmixer}
\end{figure}

The vanilla All-MLP decoder employed in earlier works~\cite{DBLP:conf/nips/XieWYAAL21} lacked adaptivity to object deformations, which weakens the token mixing of panoramic data. A novel deformable token mixer, the DMLPv2, was proposed by Zhang etal~\cite{DBLP:journals/corr/abs-2207-11860} and is demonstrated to be effective and lightweight for both spatial and channel-wise token mixing. We leverage the DMLPv2 token mixer approach at each $l$-th level of our framework, as depicted in~\cref{fig:dmlpv2tokenmixer}, which is denoted as:

\begin{align}
    \hat{\textbf{f}}_{l} &= \textbf{DPE}(\textbf{f}^{~fused}_{l}) \\
    \hat{\textbf{f}}_{l} &= \textbf{PX}(\hat{\textbf{f}}_{l}) + \textbf{CX}(\hat{\textbf{f}}_{l}) \\
    \hat{\textbf{f}}_{l} &= \textbf{DMLP}(\hat{\textbf{f}}_{l}) + \textbf{CX}(\hat{\textbf{f}}_{l}) \\
    \textbf{f}^{~decoded}_{l} &= \textbf{UpSample}(\hat{\textbf{f}}_{l})
\label{eqn:decoder}
\end{align}

The Channel Mixer (CX) of the DMLPv2 considers space-consistent yet channel-wise feature reweighting, strengthening the feature by emphasizing informative channels. Focusing on spatial-wise sampling using fixed and adaptive offsets, respectively, the Pooling Mixer (PX) and Deformable MLP (DMLP) are used in DMLPv2. The non-parametric Pooling Mixer (PX) is implemented by an average pooling operator. The adaptive data-dependent spatial offset $\boldsymbol{\Delta}^{DMLP}_{( i, j, c )}$ is predicted channel-wise.

Finally, to output the prediction for $N_{class}$ semantics masks, the decoded features from the four steps are concatenated and given to a segmentation header module, depicted in~\cref{fig:segmentationteaser}.

%------------------------------------------------------------------------
\section{Experiments}
\label{sec:experiments}

\subsection{Datasets}

For the purpose of evaluating our suggested cross-modal framework for interior settings, we use three multi-modal equirectangular semantic segmentation datasets. In each of our tests, we resize the input image to $512 \times 1024$, and then we compute evaluation metrics, such as Mean Region Intersection Over Union (mIoU), Pixel Accuracy (aAcc), and Mean Accuracy (mAcc), using the MMSegmentation IoU script\footnote{https://mmsegmentation.readthedocs.io/en/0.x/}.

\textbf{Stanford2D3DS dataset}~\cite{DBLP:journals/corr/ArmeniSZS17} contains $1713$ multi-modal equirectangular images with $13$ object categories. We split the data from area\_$1$ to area\_$6$ for training and validation in a manner similar to Armeni \etal~\cite{DBLP:journals/corr/ArmeniSZS17}, using a 3-fold cross-validation scheme, and we give the mean values across the folds. Furthermore, the publicly accessible code\footnote{https://github.com/charlesCXK/Depth2HHA-python} is used to compute the panoramic HHA~\cite{DBLP:conf/eccv/GuptaGAM14} modality using the appropriate depth and camera parameters.

\textbf{Structured3D dataset}~\cite{DBLP:conf/eccv/ZhengZLTGZ20} offers $40$ NYU-Depth-v2~\cite{DBLP:conf/eccv/SilbermanHKF12} object categories, $196515$ synthetic, multi-modal, equirectangular images with a variety of lighting setups. In line with Zheng \etal~\cite{DBLP:conf/eccv/ZhengZLTGZ20}, we establish typical training, validation, and test splits as follows: scene\_00000 to scene\_02999 for training, scene\_03000 to scene\_03249 for validation, and scene\_03250 to scene\_03499 for testing. For all of the tests we conduct, we use rendered raw lighting images with full furniture arrangements.

\textbf{Matterport3D dataset}~\cite{DBLP:conf/3dim/ChangDFHNSSZZ17} The $10800$ panoramic views in the Matterport3D~\cite{DBLP:conf/3dim/ChangDFHNSSZZ17} collection are represented by $18$ viewpoints per image frame, necessitating an explicit conversion to an equirectangular format. Second, the associated semantic annotations are spread among four files (xxx.house, xxx.ply, xxx.fsegs.json, and xxx.semseg.json). We employ the open-source matterport\_utils\footnote{https://github.com/atlantis-ar/matterport\_utils} code for post-processing, where the \textit{mpview} script is used to produce annotation images and the \textit{preparepano} script is used to stitch the $18$ images that were taken into a $360$-degree panorama. For our trials using the $40$ object categories, we created own training, validation, and test splits, refer to appendix.

\subsection{Implementation Details}

With an initial learning rate of $6$e-$5$ programmed by the poly strategy with power $0.9$ over the training epochs, we train our models using a pre-trained SegFormer MiT-B2\footnote{https://github.com/huaaaliu/RGBX\_Semantic\_Segmentation} RGB backbone on the RTXA6000 GPU. For Stanford2D3DS~\cite{DBLP:journals/corr/ArmeniSZS17}, Structured3D~\cite{DBLP:conf/eccv/ZhengZLTGZ20}, and Matterport3D~\cite{DBLP:conf/3dim/ChangDFHNSSZZ17} experiments, there are $200$ training epochs, $50$, and $100$ respectively. The optimizer AdamW~\cite{DBLP:journals/corr/KingmaB14} is employed with the following parameters: batch size $4$, epsilon $1$e-$8$, weight decay $1$e-$2$, and betas $(0.9, 0.999)$. Random horizontal flipping, random scaling to scales of $\{0.5, 0.75, 1, 1.25, 1.5, 1.75\}$, and random cropping to $512 \times 512$ are added for image argumentations. Deformable Patch Embedding module (DPE), refer to~\cref{sec:encoding}, is used for the panoramic encoder stage-1 and a conventional Overlapping Patch Embedding (OPE) module~\cite{DBLP:conf/nips/XieWYAAL21}, for the other stages of our framework. More specific settings are described in detail in the appendix.

We conducted our tests for the following fusion configurations: \textbf{RGB}-only, \textbf{RGB}-\textbf{D}epth, \textbf{RGB}-\textbf{N}ormal, \textbf{RGB}-\textbf{H}HA, and \textbf{RGB}-\textbf{D}epth-\textbf{N}ormal, \textbf{RGB}-\textbf{D}epth-\textbf{H}HA, and \textbf{RGB}-\textbf{N}ormal-\textbf{H}HA. In our tests, we only use pathways and modules in our encoding-decoding stages and skip any unnecessary parts of our framework based on these combinations. For example, in the CM-FRM and FFM modules discussed in~\cref{sec:encoding}, we employ bi-directional features for cross-modal interactions for the \textbf{RGB}-\textbf{D}epth scenario, whereas for the \textbf{RGB}-\textbf{D}epth-\textbf{N}ormal situation, we use routes that lead to tri-directional interactions across the features.

\subsection{Experiment Results and Analysis}

We carry out comprehensive tests on multimodal segmentation datasets for indoor settings to demonstrate the effectiveness of our proposed architecture of cross-modal fusion using panoramas. We employ the aforementioned training epochs, random crop-size, and batch size variables to compare our method against the current state-of-the-art approaches Trans4PASS+~\cite{DBLP:journals/corr/abs-2207-11860}, HoHoNet~\cite{DBLP:conf/cvpr/0004SC21}, PanoFormer~\cite{DBLP:conf/eccv/ShenLLNZZ22}, CMNeXt~\cite{DBLP:journals/corr/abs-2303-01480}, and TokenFusion~\cite{DBLP:conf/cvpr/00010C00022}. For a detailed description of their implementation, see the corresponding works. While all other approaches have been reproduced using the conditions of our experiment, the CBFC~\cite{DBLP:conf/wacv/ZhengLNLSZ23} and Tangent~\cite{DBLP:conf/cvpr/EderSLF20} results described here are from the related original paper. In ~\Cref{fig:semgnetationresultsbarchart},~\Cref{fig:semgnetationresultsvisualize} and~\Cref{fig:semgnetationfailurevisualize}, as well as in ~\Cref{tab:stanford2d3ds} and~\Cref{tab:structured3dmatterport3d}, are visualizations of the quantitative results and comparisons to the state-of-the-art.

\begin{table}
  \centering
  \begin{tabular}{@{}lccc@{}}
    \toprule
    \multirow{2}{*}{\textbf{Method}}                     & \multirow{2}{*}{\textbf{Modal}}   & \multicolumn{2}{c}{\textbf{3-fold Val.}} \\
                                                         &            & \textbf{mIoU (\%)}   & \textbf{mAcc (\%)} \\
    \midrule
    \midrule
    % Trans4PASS~\cite{DBLP:conf/cvpr/00010MRPS22}         & RGB        & \\
    Trans4PASS+~\cite{DBLP:journals/corr/abs-2207-11860} & \multirow{6}{*}{RGB}   & $52.04$              & $63.98$ \\
    HoHoNet~\cite{DBLP:conf/cvpr/0004SC21}               &                        & $51.99$              & $62.97$ \\
    PanoFormer~\cite{DBLP:conf/eccv/ShenLLNZZ22}         &                        & $52.35$              & $64.31$ \\
    CBFC~\cite{DBLP:conf/wacv/ZhengLNLSZ23}              &                        & $52.20$              & $65.60$ \\
    Tangent~\cite{DBLP:conf/cvpr/EderSLF20}              &                        & $45.60$              & $65.20$ \\
    \textbf{\textit{OURS}}                               &                        & $52.87$              & $63.96$ \\
    \midrule
    HoHoNet~\cite{DBLP:conf/cvpr/0004SC21}               & \multirow{4}{*}{RGB-D} & $56.73$              & $68.23$ \\
    PanoFormer~\cite{DBLP:conf/eccv/ShenLLNZZ22}         &                        & $57.03$              & $68.08$ \\
    CBFC~\cite{DBLP:conf/wacv/ZhengLNLSZ23}              &                        & $56.70$              & $70.80$ \\
    Tangent~\cite{DBLP:conf/cvpr/EderSLF20}              &                        & $52.50$              & $70.10$ \\
    \textbf{\textit{OURS}}                               &                        & $55.49$              & $68.57$ \\
    \midrule
    \multirow{5}{*}{\textbf{\textit{OURS}}}              & RGB-N                  & $58.24$              & $68.79$ \\
                                                         & RGB-H                  & $\boldsymbol{60.60}$ & $\boldsymbol{70.68}$ \\
                                                         & RGB-D-N                & $59.43$              & $69.03$ \\
                                                         & RGB-D-H                & $59.99$              & $70.44$ \\
                                                         & RGB-N-H                & $60.24$              & $70.61$ \\
    \bottomrule
  \end{tabular}
  \caption{Results on Stanford2D3DS~\cite{DBLP:journals/corr/ArmeniSZS17}.}
  \label{tab:stanford2d3ds}
\end{table}

\textbf{Results on Stanford2D3DS}~\Cref{tab:stanford2d3ds} presents the thorough comparisons between our method and other current panoramic methods. Overall, our method delivers cutting-edge performance in the merging of complementary modalities for semantic segmentation. Our method produces results that are comparable to those of existing methods~\cite{DBLP:conf/cvpr/0004SC21, DBLP:conf/eccv/ShenLLNZZ22, DBLP:conf/wacv/ZhengLNLSZ23, DBLP:conf/cvpr/EderSLF20} when used with RGB-Depth panoramas, and it further improved the results when \textbf{RGB}, \textbf{D}epth, \textbf{N}ormals, and \textbf{H}HA combinations were combined. With \textbf{RGB}-\textbf{H}HA image-based fusion, the highest mIoU was reached at $60.60\%$. By utilizing the complementary geometric, disparity, and textural information, the mIoU metric increased from \textbf{RGB}-only to gradually fusing \textbf{D}epth and \textbf{N}ormals, $52.87\% \rightarrow 55.49\% \rightarrow 59.43\%$.

\begin{table*}
  \centering
  \begin{tabular}{@{}lccccc@{}}
    \toprule
    \multirow{2}{*}{\textbf{Method}}                     & \multirow{2}{*}{\textbf{Modal}}   & \multicolumn{2}{c}{\textbf{Structured3D}}   & \multicolumn{2}{c}{\textbf{Matterport3D}} \\
                                                         &            & \textbf{Validation mIoU (\%)} & \textbf{Test mIoU (\%)} & \textbf{Validation mIoU (\%)} & \textbf{Test mIoU (\%)} \\
    \midrule
    \midrule
    % Trans4PASS~\cite{DBLP:conf/cvpr/00010MRPS22}         & RGB        &  &  & &\\
    Trans4PASS+~\cite{DBLP:journals/corr/abs-2207-11860} & \multirow{4}{*}{RGB}   & $66.74$              & $66.90$               & $33.43$              & $29.19$ \\
    HoHoNet~\cite{DBLP:conf/cvpr/0004SC21}               &                        & $66.09$              & $64.41$               & $31.91$              & $29.33$ \\
    PanoFormer~\cite{DBLP:conf/eccv/ShenLLNZZ22}         &                        & $55.57$              & $54.87$               & $30.04$              & $26.87$ \\
    \textbf{\textit{OURS}}                               &                        & $71.94$              & $68.34$               & $35.15$               & $31.30$ \\
    \midrule
    HoHoNet~\cite{DBLP:conf/cvpr/0004SC21}               & \multirow{3}{*}{RGB-D} & $69.51$              & $66.99$               & $35.36$              & $32.02$ \\
    PanoFormer~\cite{DBLP:conf/eccv/ShenLLNZZ22}         &                        & $60.98$              & $59.27$               & $33.99$              & $31.23$ \\
    \textbf{\textit{OURS}}                               &                        & $73.78$              & $70.17$              & $39.19$               & $\boldsymbol{35.92}$ \\
    \midrule
    \multirow{2}{*}{\textbf{\textit{OURS}}}              & RGB-N                  & $74.38$              & $71.00$              & $38.91$               & $35.77$ \\
                                                         & RGB-D-N                & $\boldsymbol{75.86}$ & $\boldsymbol{71.97}$ & $\boldsymbol{39.26}$ & $35.52$ \\
    \bottomrule
  \end{tabular}
  \caption{Results on Structured3D~\cite{DBLP:conf/eccv/ZhengZLTGZ20} and Matterport3D~\cite{DBLP:conf/3dim/ChangDFHNSSZZ17} datasets.}
  \label{tab:structured3dmatterport3d}
\end{table*}

\textbf{Results on Structured3D} We further test Structured3D using simply \textbf{RGB}, \textbf{D}epth, and \textbf{N}ormals, as seen in~\Cref{tab:structured3dmatterport3d}. On the validation and test data splits, our \textbf{RGB}-only model performs at the cutting edge at $71.94\%$ and $68.34\%$, respectively. Additionally, by combining depth and normals data, we were able to outperform benchmark results for (validation, test) by $(+1.84, +1.83)$ for \textbf{RGB}-\textbf{D}epth, $(+2.44, +2.66)$ for \textbf{RGB}-\textbf{N}ormals, and $(+3.92, 3.63)$ for \textbf{RGB}-\textbf{D}epth-\textbf{N}ormals fusion.

\textbf{Results on Matterport3D}~\Cref{tab:structured3dmatterport3d} shows further trials using Matterport3D~\cite{DBLP:conf/3dim/ChangDFHNSSZZ17} with comparable \textbf{RGB}, \textbf{D}epth, and \textbf{N}ormals combinations in addition to the Structured3D~\cite{DBLP:conf/eccv/ZhengZLTGZ20} dataset. Our method outperforms the current panoramic techniques in this case for both \textbf{RGB}-only and \textbf{RGB}-\textbf{D}epth based semantic segmentation. Our validation and test pair mIoU metrics values for \textbf{RGB}-only and \textbf{RGB}-\textbf{D}epth, respectively, are $(35.15\%, 31.30\%)$ and $(39.19\%, 35.92\%)$, respectively, when compared to the benchmark. However, we discovered that the combination of the multi-modal fusion with normals did not result in the expected improvement in performance, as demonstrated in other tests, $(38.91\%, 35.92\%)$ for \textbf{RGB}-\textbf{N}ormal and $(39.26\%, 35.52\%)$ for \textbf{RGB}-\textbf{D}epth-\textbf{N}ormal. Our hypothesis is that the depth and normals data result in a limited amount of modal differences, and thus modal addition may be unnecessary.

\subsection{Qualitative Analysis}

\begin{figure*}
  \centering
  \includegraphics[width=\linewidth]{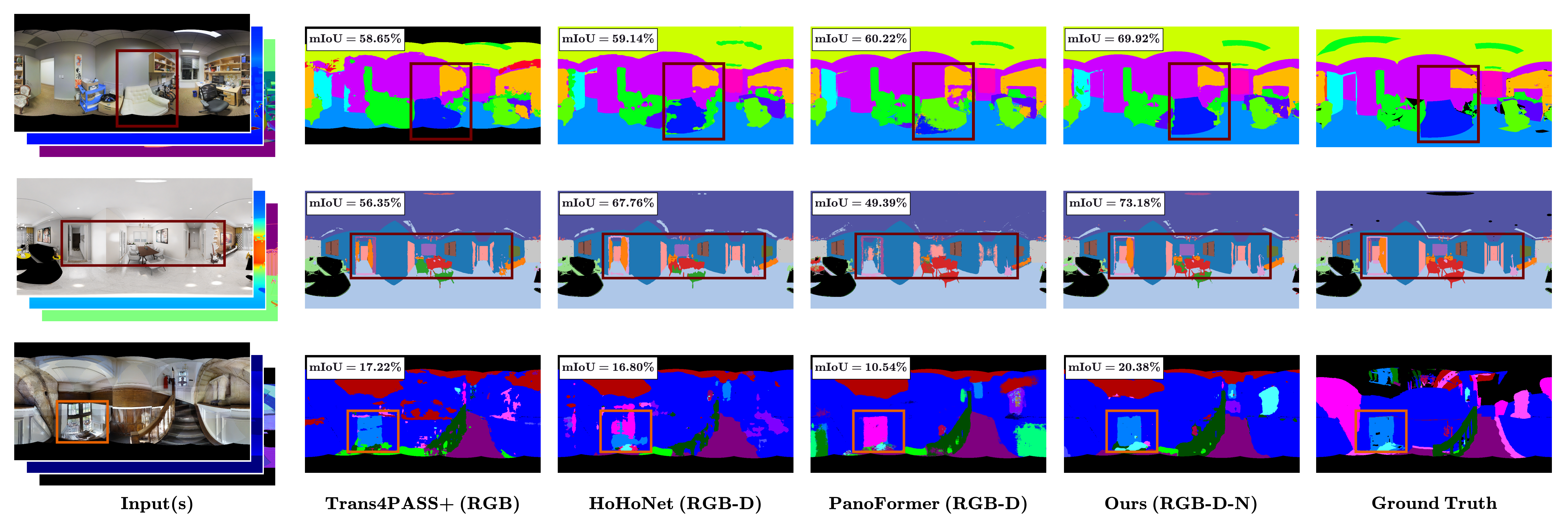}

  \caption{Results of multi-modal panoramic semantic segmentation for the \textbf{RGB}-only, \textbf{RGB}-\textbf{D}epth, and \textbf{RGB}-\textbf{D}epth-\textbf{N}ormals methods are visualized. For \textbf{RGB} segmentation, we use Trans4PASS+~\cite{DBLP:journals/corr/abs-2207-11860} baseline, which employs the same SegFormer MiT-B2 backbone~\cite{DBLP:conf/nips/XieWYAAL21} with Deformable Patch Embeddings (DPE) and DMLPv2 decoder as ours, as detailed in~\cref{sec:decoding}. PanoFormer~\cite{DBLP:conf/eccv/ShenLLNZZ22} uses a cutting-edge panoramic transformer-based architecture for \textbf{RGB}-\textbf{D}epth segmentation, while HoHoNet~\cite{DBLP:conf/cvpr/0004SC21} is built on pre-trained ResNet-101~\cite{DBLP:conf/cvpr/HeZRS16} in conjunction with a sophisticated horizon-to-dense module. Our strategy leverages \textbf{RGB}-\textbf{D}epth-\textbf{N}ormal fusion to improve performance by utilizing all available features.}
   \label{fig:semgnetationresultsvisualize}
\end{figure*}

The segmentation outcomes of panoramic techniques are shown in~\cref{fig:semgnetationresultsvisualize}, which displays the findings from left to right and from top to bottom across several indoor datasets. Overall, our approach is able to take advantage of depth and geometry data as well as textures from \textbf{RGB}, \textbf{D}epth and \textbf{N}ormal modalities and correctly identify object semantics with a better level of accuracy, as indicated. While our baseline Trans4PASS+~\cite{DBLP:journals/corr/abs-2207-11860} accurately classifies the book shelf, sofa, and chair in the first row, the architecture was unable to predict the exact geometrical shapes. Using depth information, PanoFormer~\cite{DBLP:conf/eccv/ShenLLNZZ22} and HoHoNet~\cite{DBLP:conf/cvpr/0004SC21} were able to estimate the exact geometry of the chair and bookshelf, however, former method incorrectly guessed the object class of the sofa. The third row findings of the \textbf{RGB}-only and \textbf{RGB}-\textbf{D}epth based techniques show a similar trend. When compared to current state-of-the-art baselines, our method consistently predicted geometric shapes that were considerably clearer and had precise object semantics in these situations. The approach can even handle thin structures like the neck of a guitar and items on a dining table, as shown in the second row.

\begin{figure*}
  \centering
  \includegraphics[width=\linewidth]{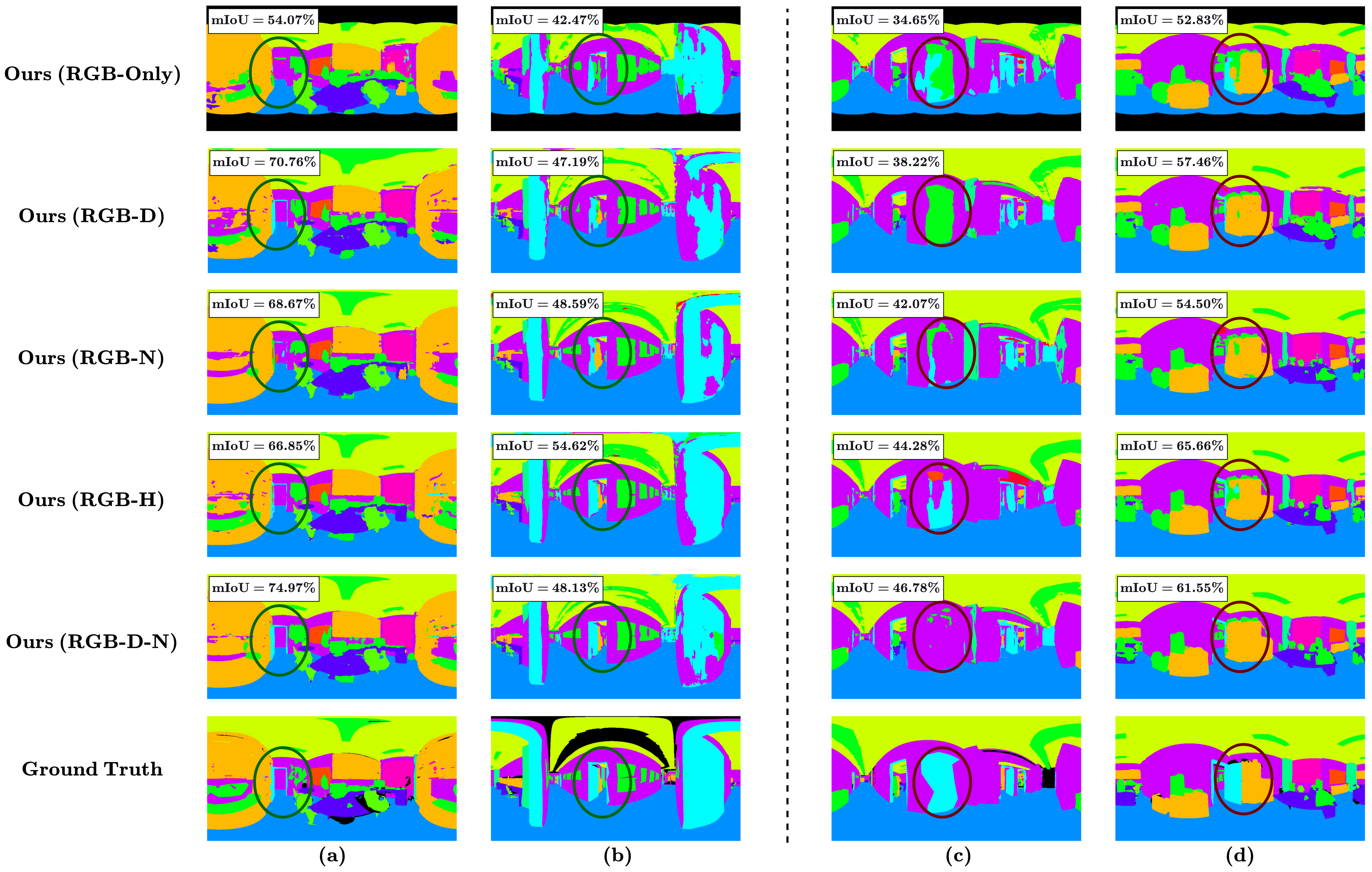}

  \caption{Visualization of semantic segmentation results for our framework using Stanford2D3DS\cite{DBLP:journals/corr/ArmeniSZS17} for \textbf{RGB}-only, \textbf{RGB}-\textbf{D}epth, \textbf{RGB}-\textbf{N}ormals, \textbf{RGB}-\textbf{H}HA, and \textbf{RGB}-\textbf{D}epth-\textbf{N}ormals (top-to-bottom) combinations. By utilizing complementary traits, our method was successful in identifying deformed and visually identical building structures like doors in columns (a) and (b). Under ambiguity, we were unable to differentiable between the distorted door and the wall or the deformed door and the bookcase in columns (c) and (d), respectively.}
   \label{fig:semgnetationfailurevisualize}
\end{figure*}

The qualitative results of different Stanford2D3DS~\cite{DBLP:journals/corr/ArmeniSZS17} multi-modal combinations, including \textbf{RGB}-only, \textbf{RGB}-\textbf{D}epth, \textbf{RGB}-\textbf{N}ormal, \textbf{RGB}-\textbf{H}HA, and \textbf{RGB}-\textbf{D}epth-\textbf{N}ormal, are shown in~\cref{fig:semgnetationfailurevisualize} using our paradigm. While in the scenarios shown in~\cref{fig:semgnetationfailurevisualize} (a) and~\cref{fig:semgnetationfailurevisualize} (b), using complementary data from other modalities is advantageous, this may not always be the case when the model cannot tell the difference between the distorted door and the wall (\cref{fig:semgnetationfailurevisualize} (c)), or the distorted door and the bookshelf (\cref{fig:semgnetationfailurevisualize} (d)). We hypothesize that these failed cases happened as a result of the scene objects' ambiguity, which makes it difficult to distinguish using any of the accessible modalities.

\subsection{Ablation Studies}

In the context of panoramic semantic segmentation, we investigated the state-of-the-art fusion architectures CMX~\cite{DBLP:journals/corr/abs-2203-04838}, CMNeXt~\cite{DBLP:journals/corr/abs-2303-01480}, and TokenFusion~\cite{DBLP:conf/cvpr/00010C00022}. Our architecture, which was expanded to include a tri-modal panoramas scenario, is inspired on CMX~\cite{DBLP:journals/corr/abs-2203-04838}. In order to address panorama distortions, Deformable Patch Embeddings (DPE) modules, which are detailed in~\cref{sec:encoding}, are added to these encoder's backbone. The stages of the panorama decoder, as defined in~\cref{sec:decoding}, have not changed. We employ two versions of CMNeXt~\cite{DBLP:journals/corr/abs-2303-01480}, one with and one without a Self-Query Hub (SQ-Hub), with the former version demonstrating the ability to handle up to $81$ modalities with minimal overhead and processing demands. Furthermore, it is expected that SQ-Hub will soft-select informative features while remaining robust to sensor failure.

\begin{table*}
  \centering
  \begin{tabular}{@{}lccccccc@{}}
    \toprule
    \multirow{2}{*}{\textbf{Method}}                                         & \multirow{2}{*}{\textbf{Modal}}             & \multicolumn{2}{c}{\textbf{Stanford2D3DS}~\cite{DBLP:journals/corr/ArmeniSZS17}}       & \multicolumn{2}{c}{\textbf{Structured3D}~\cite{DBLP:conf/eccv/ZhengZLTGZ20}}    & \multicolumn{2}{c}{\textbf{Matterport3D}~\cite{DBLP:conf/3dim/ChangDFHNSSZZ17}} \\
            &   & \textbf{mIoU (\%)}   & \textbf{mAcc (\%)} & \textbf{mIoU (\%)}   & \textbf{mAcc (\%)} & \textbf{mIoU (\%)}   & \textbf{mAcc (\%)}\\
    \midrule
    \midrule
    {\textit{OURS}} - TokenFusion~\cite{DBLP:conf/cvpr/00010C00022}       & \multirow{4}{*}{RGB-D}     & $58.88$              & $68.57$       & $62.58$  &     $70.54$   & $\boldsymbol{36.48}$     & $49.30$ \\
    {\textit{OURS}} - CMNeXt (S)~\cite{DBLP:journals/corr/abs-2303-01480} &                            & $56.49$              & $66.27$       & $68.35$  &     $76.54$   & $35.38$                  & $49.71$ \\
    {\textit{OURS}} - CMNeXt    ~\cite{DBLP:journals/corr/abs-2303-01480} &                            & $54.27$              & $64.13$       & $69.31$  &     $78.12$   & $34.99$                  & $49.42$ \\
    {\textit{OURS}}                                                       &                            & $55.49$              & $66.02$       & $70.17$  &     $77.88$   & $35.92$                  & $49.24$ \\
    \midrule
    {\textit{OURS}} - TokenFusion~\cite{DBLP:conf/cvpr/00010C00022}       & \multirow{4}{*}{RGB-N}     & $57.86$              & $67.39$       & $62.76$  &     $70.91$   & $35.71$                  & $48.92$ \\
    {\textit{OURS}} - CMNeXt (S)~\cite{DBLP:journals/corr/abs-2303-01480} &                            & $53.61$              & $63.26$       & $68.47$   &    $76.82$   & $33.10$                  & $46.32$ \\
    {\textit{OURS}} - CMNeXt    ~\cite{DBLP:journals/corr/abs-2303-01480} &                            & $50.47$              & $60.83$       & $68.62$   &    $76.99$   & $33.80$                  & $47.02$ \\
    {\textit{OURS}}                                                       &                            & $58.24$              & $68.79$       & $71.00$   &    $78.68$   & $35.77$                  & $\boldsymbol{50.39}$ \\
    \midrule
    {\textit{OURS}} - TokenFusion~\cite{DBLP:conf/cvpr/00010C00022}       & \multirow{4}{*}{RGB-H}     & $59.06$              & $68.07$       & $-$         &    $-$         & $-$                        & $-$ \\
    {\textit{OURS}} - CMNeXt (S)~\cite{DBLP:journals/corr/abs-2303-01480} &                            & $55.70$              & $65.79$       & $-$         &    $-$         & $-$                        & $-$ \\
    {\textit{OURS}} - CMNeXt    ~\cite{DBLP:journals/corr/abs-2303-01480} &                            & $52.48$              & $62.78$       & $-$         &    $-$         & $-$                        & $-$ \\
    {\textit{OURS}}                                                       &                            & $\boldsymbol{60.60}$ & $\boldsymbol{70.68}$         & $-$         &    $-$       & $-$                        & $-$ \\
    \midrule
    {\textit{OURS}} - CMNeXt (S)~\cite{DBLP:journals/corr/abs-2303-01480} & \multirow{3}{*}{RGB-D-H}   & $57.62$              & $67.80$       & $-$         &    $-$         & $-$                        & $-$ \\
    {\textit{OURS}} - CMNeXt    ~\cite{DBLP:journals/corr/abs-2303-01480} &                            & $54.54$              & $64.22$       & $-$         &    $-$         & $-$                        & $-$ \\
    {\textit{OURS}}                                                       &                            & $59.99$              & $70.44$       & $-$         &    $-$         & $-$                        & $-$ \\
    \midrule
    {\textit{OURS}} - CMNeXt (S)~\cite{DBLP:journals/corr/abs-2303-01480} & \multirow{3}{*}{RGB-D-N}   & $55.72$              & $65.86$       & $69.55$     &    $77.50$     & $35.18$                    & $49.79$ \\
    {\textit{OURS}} - CMNeXt    ~\cite{DBLP:journals/corr/abs-2303-01480} &                            & $54.65$              & $64.53$       & $69.11$     &    $77.54$     & $35.55$                    & $50.09$ \\
    {\textit{OURS}}                                                       &                            & $59.43$              & $69.03$       & $\boldsymbol{71.97}$     & $\boldsymbol{79.67}$      & $35.52$   & $50.01$ \\
    \midrule
    {\textit{OURS}} - CMNeXt (S)~\cite{DBLP:journals/corr/abs-2303-01480} & \multirow{3}{*}{RGB-N-H}   & $55.45$              & $65.24$       & $-$         &    $-$         & $-$                        & $-$ \\
    {\textit{OURS}} - CMNeXt    ~\cite{DBLP:journals/corr/abs-2303-01480} &                            & $52.50$              & $62.19$       & $-$         &    $-$         & $-$                        & $-$ \\
    {\textit{OURS}}                                                       &                            & $60.24$              & $70.62$       & $-$         &    $-$         & $-$                        & $-$ \\
    \midrule
    {\textit{OURS}} - CMNeXt (S)~\cite{DBLP:journals/corr/abs-2303-01480} & \multirow{2}{*}{RGB-D-N-H} & $55.55$              & $65.33$       & $-$         &    $-$         & $-$                        & $-$ \\
    {\textit{OURS}} - CMNeXt    ~\cite{DBLP:journals/corr/abs-2303-01480} &                            & $54.48$              & $64.21$       & $-$         &    $-$         & $-$                        & $-$ \\
    \bottomrule
  \end{tabular}
  \caption{An analysis of the various cross-modal fusion techniques applied to the encoder stages of our multi-modal panoramic architecture.}
  \label{tab:fusionablation}
\end{table*}

~\Cref{tab:fusionablation} compares $\{$ \textbf{RGB}-\textbf{D}epth, \textbf{RGB}-\textbf{N}ormals, and \textbf{RGB}-\textbf{H}HA  $\}$ bi-modal fusion, $\{$ \textbf{RGB}-\textbf{D}epth-\textbf{N}ormal, \textbf{RGB}-\textbf{D}epth-\textbf{H}HA, and \textbf{RGB}-\textbf{N}ormal-\textbf{H}HA $\}$ tri-modal fusion, and $\{$ \textbf{RGB}-\textbf{D}epth-\textbf{N}ormal-\textbf{H}HA $\}$ quad-modal fusion.  Overall, the CMX~\cite{DBLP:journals/corr/abs-2203-04838} technique we adopted had greater performance. Our methodology, which uses TokenFusion~\cite{DBLP:conf/cvpr/00010C00022} for feature extraction and fusion, performs well on the Matterport3D~\cite{DBLP:conf/3dim/ChangDFHNSSZZ17} dataset, although it lags behind Stanford3D2DS~\cite{DBLP:journals/corr/ArmeniSZS17} and Structured3D\cite{DBLP:conf/eccv/ZhengZLTGZ20} by a wider margin. Thanks to Self-Query Hub (SQ-Hub), our approach to using encoded features from CMNeXt~\cite{DBLP:journals/corr/abs-2303-01480} performs comparably across datasets with fewer computational overload. However, in the majority of cases, in our panoramic trials, we have observed similar outcomes without SQ-Hub.

%------------------------------------------------------------------------
\section{Conclusion}
\label{sec:conclusion}

In this work, we revisit multi-modal semantic segmentation at the pixel level for a holistic scene understating. Through a cutting-edge panoramic encoder design, we present the framework with distortion awareness and cross-modal interactions. Our encoder learns severe object deformations and panoramic image distortions with equirectangular representations, and leverages feature interaction and feature fusion for cross-modal global reasoning in RGB-X panoramic segmentation. Our architecture produces superior performance on indoor panoramic benchmarks using RGB-Depth, RGB-Normal, and RGB-HHA combinations. Furthermore, we rebuild our cross-modal panoramic encoder to learn textual, disparity, and geometrical features using tri-modal (RGB-Depth-Normals) fusion, hence removing the requirement to compute HHA representations while maintaining the same performance. One major drawback of our method is that having two or more input streams active at once typically results in a large rise in complexity, refer to appendix. We'll look for techniques to combine multi-modal panoramas and 3D LiDAR data in the future with the least amount of processing effort possible.

\begin{flushleft}
\textbf{Acknowledgement.} This work was partially funded by the EU Horizon Europe Framework Program under grant agreement 101058236 (HumanTech).
\end{flushleft}

%%%%%%%%% APPENDIX
\appendix

%%%%%%%%% BODY TEXT - ENTER YOUR RESPONSE BELOW
\section{Experimentation details}

\subsection{Matterport3D dataset}

To divide the $10800$ panoramic equirectangular images in the Matterport3D~\cite{DBLP:conf/3dim/ChangDFHNSSZZ17} dataset, we create standard training, evaluation, and test splits. The $90$ building-scale scenarios, which included a range of scene types like residences, offices, and churches, were divided into an $80$-$10$-$10$ split. For all our segmentation experiments using the $40$ object categories, we use these training, validation, and test splits.

\begin{table}
  \centering
  \footnotesize
  \begin{tabular}{@{}ll@{}}
    \toprule
    \textbf{Split} & \textbf{Building Scene Ids}\\
    \midrule
    \midrule
    Evaluation & \texttt{UwV83HsGsw3, X7HyMhZNoso, Z6MFQCViBuw, } \\
               & \texttt{e9zR4mvMWw7, q9vSo1VnCiC, rPc6DW4iMge, } \\
               & \texttt{rqfALeAoiTq, uNb9QFRL6hY, wc2JMjhGNzB, } \\
               & \texttt{x8F5xyUWy9e, yqstnuAEVhm} \\
    \midrule
    Testing    & \texttt{VFuaQ6m2Qom, VLzqgDo317F, ZMojNkEp431, } \\
               & \texttt{jh4fc5c5qoQ, jtcxE69GiFV, pRbA3pwrgk9, } \\
               & \texttt{pa4otMbVnkk, D7G3Y4RVNrH, dhjEzFoUFzH, } \\
               & \texttt{GdvgFV5R1Z5, gYvKGZ5eRqb, YmJkqBEsHnH, } \\
    \midrule
    Training   & \texttt{/* all other scenes excluded from } \\
               & \texttt{   evaluation \& testing splits */} \\
%    Training   & \texttt{17DRP5sb8fy, 1LXtFkjw3qL, 1pXnuDYAj8r, } \\
%               & \texttt{29hnd4uzFmX, 2azQ1b91cZZ, 2n8kARJN3HM, } \\
%               & \texttt{2t7WUuJeko7, 5LpN3gDmAk7, 5q7pvUzZiYa, } \\
%               & \texttt{5ZKStnWn8Zo, 759xd9YjKW5, 7y3sRwLe3Va, } \\
%               & \texttt{8194nk5LbLH, 82sE5b5pLXE, 8WUmhLawc2A, } \\
%               & \texttt{aayBHfsNo7d, ac26ZMwG7aT, ARNzJeq3xxb, } \\
%               & \texttt{B6ByNegPMKs, b8cTxDM8gDG, cV4RVeZvu5T, } \\
%               & \texttt{D7N2EKCX4Sj, E9uDoFAP3SH, EDJbREhghzL, } \\
%               & \texttt{EU6Fwq7SyZv, fzynW3qQPVF, gTV8FGcVJC9, } \\
%               & \texttt{gxdoqLR6rwA, gZ6f7yhEvPG, HxpKQynjfin, } \\
%               & \texttt{i5noydFURQK, JeFG25nYj2p, JF19kD82Mey, } \\
%               & \texttt{JmbYfDe2QKZ, kEZ7cmS4wCh, mJXqzFtmKg4, } \\
%               & \texttt{oLBMNvg9in8, p5wJjkQkbXX, pLe4wQe7qrG, } \\
%               & \texttt{Pm6F8kyY3z2, PuKPg4mmafe, PX4nDJXEHrG, } \\
%               & \texttt{qoiz87JEwZ2, QUCTc6BB5sX, r1Q1Z4BcV1o, } \\
%               & \texttt{r47D5H71a5s, RPmz2sHmrrY, s8pcmisQ38h, } \\
%               & \texttt{S9hNv5qa7GM, sKLMLpTHeUy, SN83YJsR3w2, } \\
%               & \texttt{sT4fr6TAbpF, TbHJrupSAjP, ULsKaCPVFJR, } \\
%               & \texttt{ur6pFq6Qu1A, Uxmj2M2itWa, V2XKFyX4ASd, } \\
%               & \texttt{Vt2qJdWjCF2, VVfe2KiqLaN, Vvot9Ly1tCj, } \\
%               & \texttt{vyrNrziPKCB, VzqfbhrpDEA, WYY7iVyf5p8, } \\
%               & \texttt{XcA2TqTSSAj, YFuZgdQ5vWj, YVUC4YcDtcY, } \\
%               & \texttt{zsNo4HB9uLZ} \\
    \bottomrule
  \end{tabular}
  \caption{Dataset split for Matterport3D~\cite{DBLP:conf/3dim/ChangDFHNSSZZ17} segmentation.}
  \label{tab:example}
\end{table}

%------------------------------------------------------------------------
\section{Qualitative analysis}

\subsection{Multi-modal panoramic semantic segmentation}

\begin{figure*}
  \centering
  \includegraphics[width=0.94\linewidth]{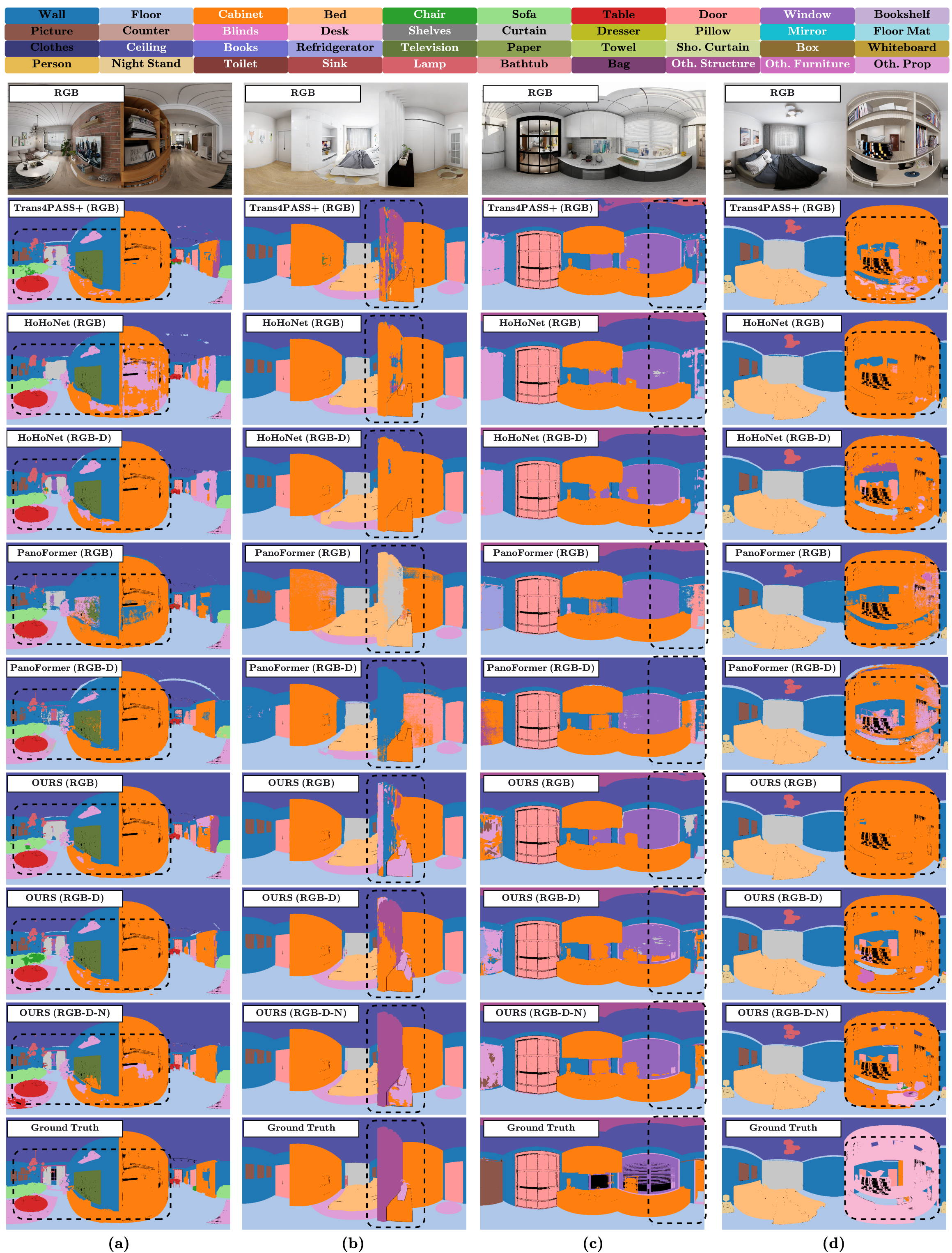}

  \caption{Structured3D~\cite{DBLP:conf/eccv/ZhengZLTGZ20} segmentation visualizations. Zoom in for better view..}
   \label{fig:structured3dresultsvisualize}
\end{figure*}

\begin{figure*}
  \centering
  \includegraphics[width=\linewidth]{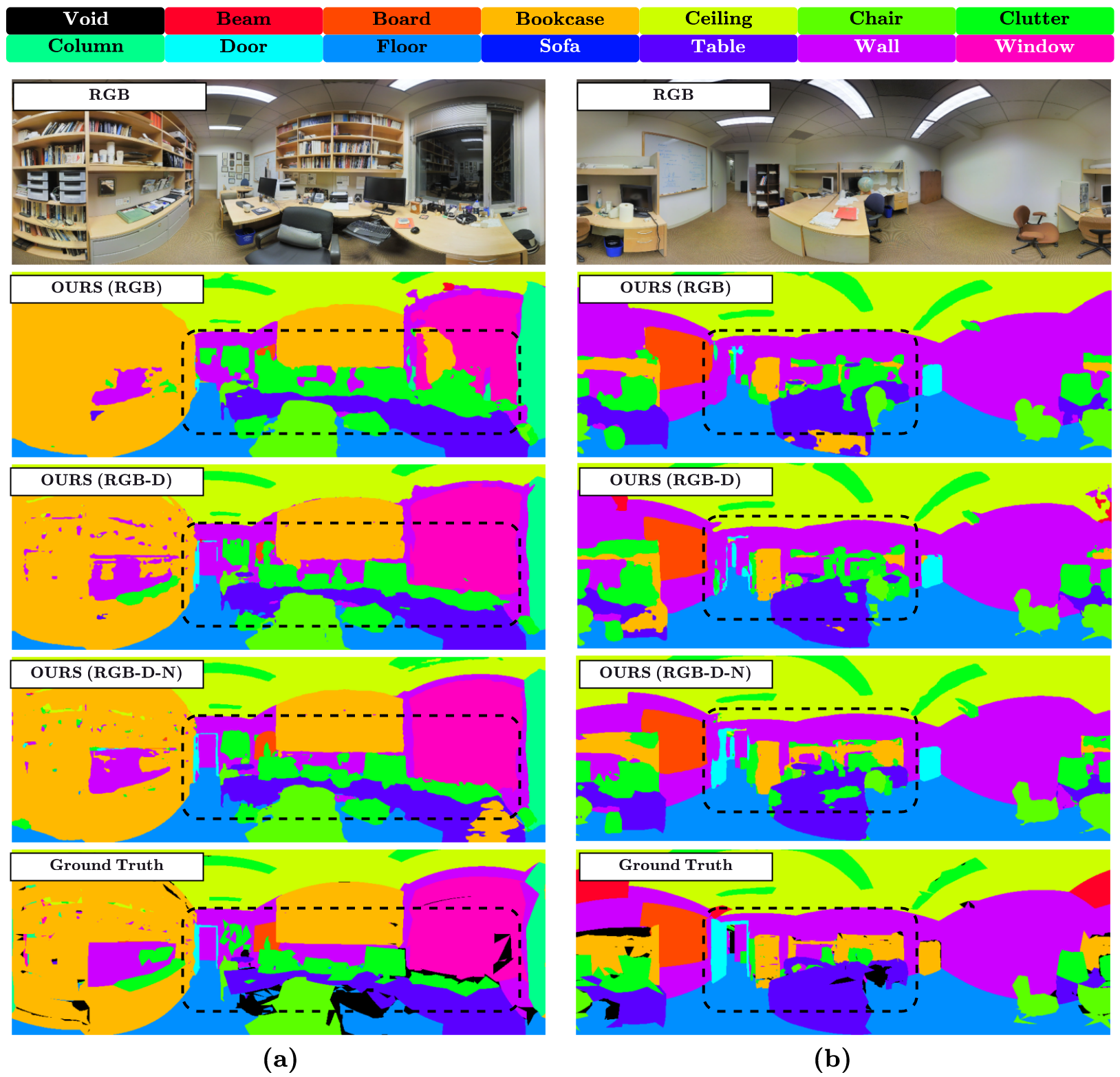}

  \caption{Stanford2D3DS~\cite{DBLP:journals/corr/ArmeniSZS17} segmentation visualizations. Zoom in for better view.}
   \label{fig:stanford2d3ds3dresultsvisualize}
\end{figure*}

\Cref{fig:stanford2d3ds3dresultsvisualize} and~\Cref{fig:structured3dresultsvisualize}, which come from the Stanford2D3DS~\cite{DBLP:journals/corr/ArmeniSZS17} evaluation set and the Structured3D~\cite{DBLP:conf/eccv/ZhengZLTGZ20} test set, respectively, show further qualitative comparisons between various fusion combinations for our proposed framework. In~\cref{fig:stanford2d3ds3dresultsvisualize} (a) and (b), our tri-model (\textbf{RGB}-\textbf{D}-\textbf{N}) is able to give better segmentation results in the categories denoted by the black dashed rectangles, such as the \textit{Door}, \textit{Window}, and \textit{Bookshelf}, while the baseline (\textbf{RGB}-only) model struggles to recognize these significantly distorted objects. The \textbf{RGB}-only baseline models wrongly segment the \textit{Door} in figure~\cref{fig:structured3dresultsvisualize} (c) as a part of the \textit{Wall}. Our tri-model (\textbf{RGB}-\textbf{D}-\textbf{N}) in this case achieves the correct segmentation results with greater accuracy than \textbf{RGB}-\textbf{D} techniques. The same conditions apply to the \textit{Cabinet} in~\cref{fig:structured3dresultsvisualize} (a) and the support between the \textit{Bed} and \textit{Cabinet} in~\cref{fig:structured3dresultsvisualize} (b). Compared to other approaches, In~\cref{fig:structured3dresultsvisualize} (d), along with the precise geometry shapes for objects placed inside the \textit{Cabinet} structure, a better segmentation result from our multi-modal (\textbf{RGB}-\textbf{D}-\textbf{N}) is displayed. However, due to visual ambiguity, the category is incorrectly predicted by all models.

%------------------------------------------------------------------------
\section{Quantitative analysis}

\subsection{Computational complexity}

For tri-modal (\textbf{RGB}-\textbf{D}epth-\textbf{N}ormals), bi-modal (\textbf{RGB}-\textbf{D}epth), and uni-modal (\textbf{RGB}-Only) panoramic fusion on Stanford2D3DS~\cite{DBLP:journals/corr/ArmeniSZS17}, we compare the computational complexity of our framework with that of existing methods in~\cref{tab:complutationalcomplexity}. As the number of input streams rises, our study indicates that our method's complexity also significantly rises.

\begin{table}
  \centering
  \begin{tabular}{@{}clccc@{}}
    \toprule
    \textbf{\#Inputs}       &  \textbf{Method}                                     & \textbf{\#Params (G)} & \textbf{TFLOPs} \\
    \midrule
    \midrule
    \multirow{4}{*}{Unary}  & Trans4PASS+~\cite{DBLP:journals/corr/abs-2207-11860} & $0.039$                & $0.131$ \\
                            & HoHoNet~\cite{DBLP:conf/cvpr/0004SC21}               & $0.070$                & $0.125$ \\
                            & PanoFormer~\cite{DBLP:conf/eccv/ShenLLNZZ22}         & $0.020$                & $0.081$ \\
                            & {\textit{OURS}}                                      & $0.040$                & $0.079$ \\
    \midrule
    \multirow{3}{*}{Binary} & HoHoNet~\cite{DBLP:conf/cvpr/0004SC21}               & $0.070$                & $0.126$ \\
                            & PanoFormer~\cite{DBLP:conf/eccv/ShenLLNZZ22}         & $0.020$                & $0.081$ \\
                            & {\textit{OURS}}                                      & $0.081$                & $0.106$ \\
    \midrule
    {Ternary}               & {\textit{OURS}}                                      & $0.123$                & $0.133$ \\
    \bottomrule
  \end{tabular}
  \caption{Comparison of computational complexity calculated @ $ 512 \times 1024 \times 3$ input dimensional.}
  \label{tab:complutationalcomplexity}
\end{table}

\subsection{Detailed results in indoor scenarios}

More qualitative comparisons based on three-fold cross validation of Stanford2D3DS\cite{DBLP:journals/corr/ArmeniSZS17} indoor scenarios are shown in~\cref{tab:stanford2d3dsperclass} to support our propose approach. When compared to the current panoramic approaches, our multi-model fusion models segment objects in regularly used categories including ceiling, wall, floor, window, and office furniture better. Our \textbf{RGB}-\textbf{D}epth-\textbf{N}ormals fusion model receives top score mIoU in $8$ out of $13$ categories. However, this model struggled to segment the \textit{Beam}, \textit{Column}, and \textit{Wall} categories.

\begin{figure*}
  \centering
  \begin{subfigure}{0.49\linewidth}
    % \fbox{\rule{0pt}{2in} \rule{.9\linewidth}{0pt}}
    \includegraphics[width=\linewidth]{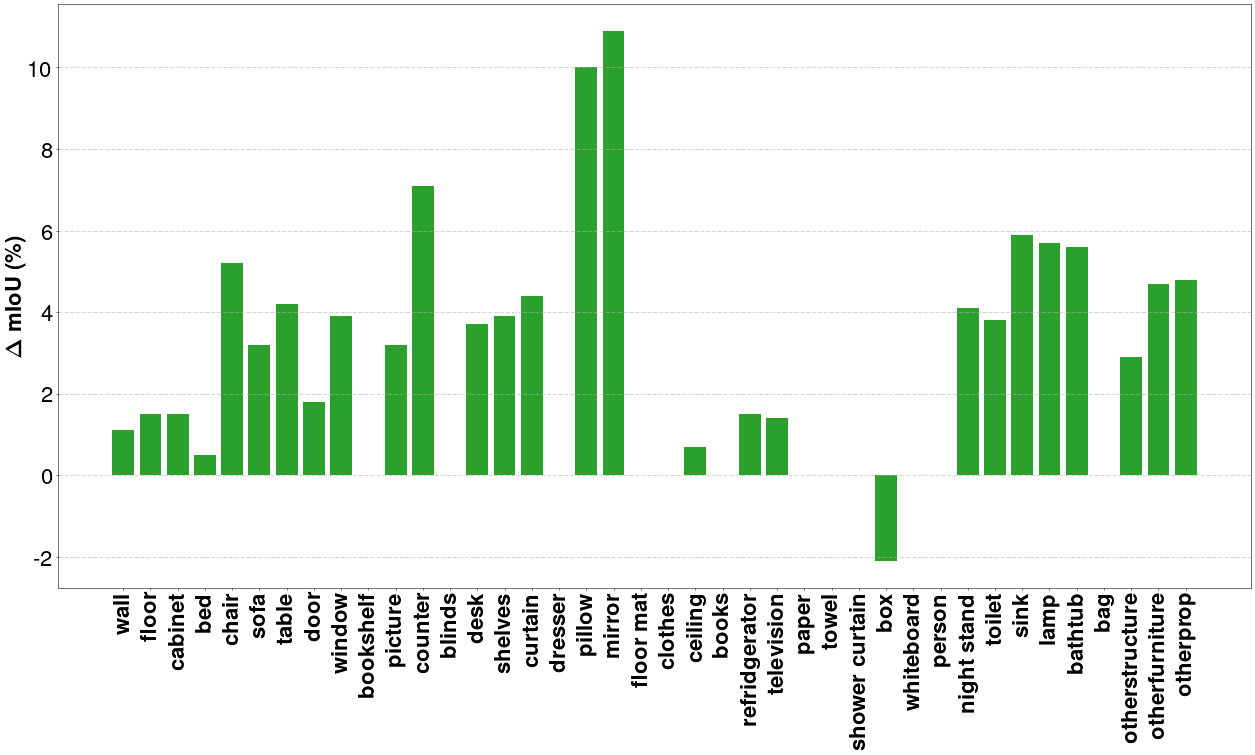}
    \caption{Structured3D~\cite{DBLP:conf/eccv/ZhengZLTGZ20}}
  \end{subfigure}
  \begin{subfigure}{0.49\linewidth}
    % \fbox{\rule{0pt}{2in} \rule{.9\linewidth}{0pt}}
    \includegraphics[width=\linewidth]{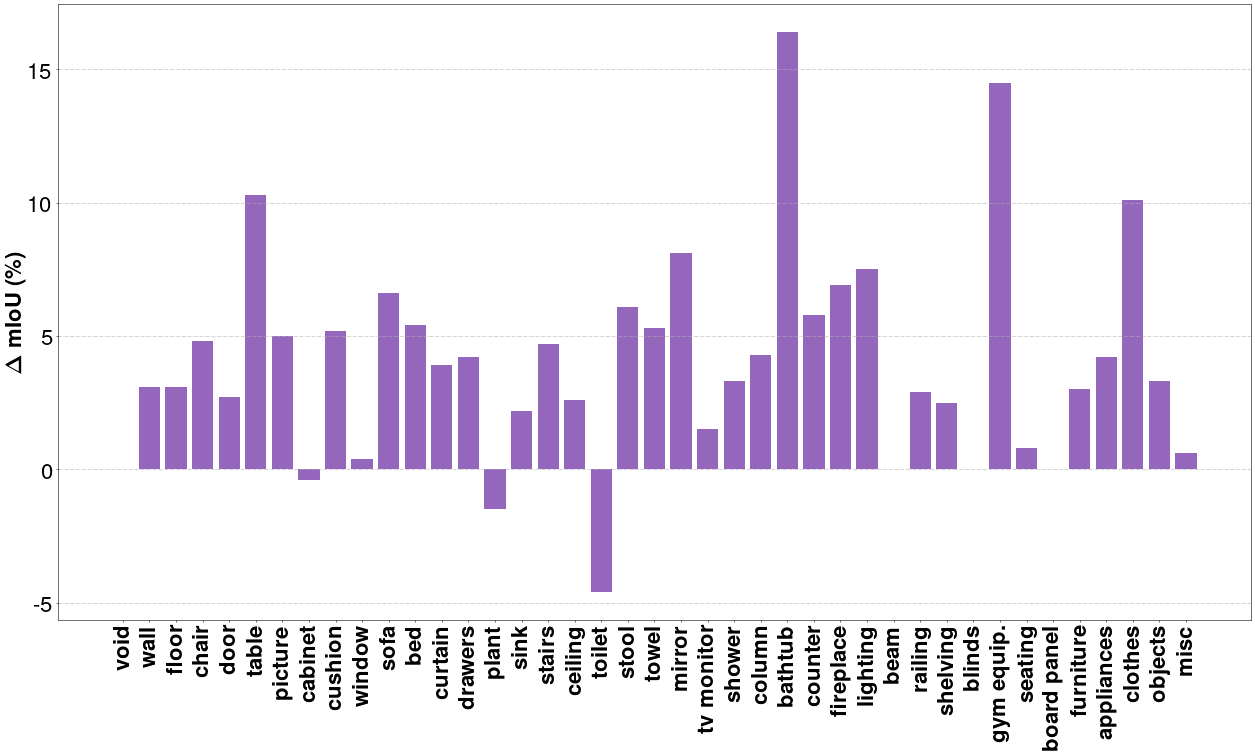}
    \caption{Matterport3D~\cite{DBLP:conf/3dim/ChangDFHNSSZZ17}}
  \end{subfigure}
  \caption{Per-class mIoU (\%) gain of OURS (\textbf{RGB}-\textbf{D}epth-\textbf{N}ormals) multi-modal panoramic semantic segmentation over baseline \textbf{RGB}-only (OURS) from Structure3D (\textit{left}) and Matterport3D (\textit{right}) test splits. Zoom in for better view.}
  \label{fig:segmentationperclassresults}
\end{figure*}

\Cref{fig:segmentationperclassresults} shows the advantage of combining multi-modalities, such as \textbf{RGB}, \textbf{D}epth, and \textbf{N}ormals, over the baseline of our technique that uses \textbf{RGB} alone to utilize complimentary textual, geometric, and disparity information. With our tri-fusion model (\textbf{RGB}-\textbf{D}-\textbf{N}), we generally observe a considerable improvement across all object categories. For the \textit{Pillow} and Mirror categories on Structured3D~\cite{DBLP:conf/eccv/ZhengZLTGZ20}, refer~\cref{fig:segmentationperclassresults} (a), as well as the \textit{Bathtub} and \textit{Gym Equipment} categories on Matterport3D~\cite{DBLP:conf/3dim/ChangDFHNSSZZ17}, refer~\cref{fig:segmentationperclassresults} (b), we saw a considerable rise of mIoU of up to $10\%$ and $15\%$, respectively. However, the box category on Structured3D~\cite{DBLP:conf/eccv/ZhengZLTGZ20} and the \textit{Cabinet}, \textit{Plant}, and \textit{Toilet} categories on ~\cite{DBLP:conf/3dim/ChangDFHNSSZZ17} also had drops of $1\%$ to $4\%$.

\begin{table*}
  \centering
  \footnotesize
  \begin{tabular}{@{}lcc|ccccccccccccc@{}}
    \toprule
    \textbf{Method} & \rotatebox[origin=c]{90}{\textbf{Modal}}   & \rotatebox[origin=c]{90}{\textbf{mIoU}}  & \rotatebox[origin=c]{90}{beam}  & \rotatebox[origin=c]{90}{board} & \rotatebox[origin=c]{90}{bookcase} & \rotatebox[origin=c]{90}{ceiling}      & \rotatebox[origin=c]{90}{chair} & \rotatebox[origin=c]{90}{clutter}   & \rotatebox[origin=c]{90}{column}   & \rotatebox[origin=c]{90}{door} & \rotatebox[origin=c]{90}{floor} & \rotatebox[origin=c]{90}{sofa}      & \rotatebox[origin=c]{90}{table}    & \rotatebox[origin=c]{90}{wall} & \rotatebox[origin=c]{90}{window} \\
    \midrule
    \midrule
    Trans4PASS+~\cite{DBLP:journals/corr/abs-2207-11860}     & \multirow{6}{*}{\rotatebox[origin=c]{90}{RGB}}   &  $52.0$ & $11.9$ & $63.2$ & $52.4$ & $81.8$ & $55.8$ & $37.4$ & $18.0$ & $59.1$ & $89.1$ & $30.0$ & $55.8$ & $70.3$ & $51.7$ \\
    HoHoNet~\cite{DBLP:conf/cvpr/0004SC21}         &                                                  & $52.0$ & $9.7$  & $61.4$ & $50.8$ & $82.3$ & $54.6$ & $35.1$ & $18.2$ & $61.3$ & $89.6$ & $34.0$ & $54.5$ & $71.7$ & $52.6$ \\
    PanoFormer~\cite{DBLP:conf/eccv/ShenLLNZZ22}      &                                                  & $52.3$ & $8.1$  & $62.1$ & $52.6$ & $83.7$ & $53.1$ & $36.9$ & $18.8$ & $64.6$ & $90.3$ & $29.4$ & $57.2$ & $72.7$ & $51.0$ \\
    CBFC~\cite{DBLP:conf/wacv/ZhengLNLSZ23}            &                                                  & $52.2$ & $-$    & $-$    & $-$    & $-$         & $-$    & $-$    & $-$    & $-$    & $-$    & $-$    & $-$    & $-$    & $-$    \\
    Tangent~\cite{DBLP:conf/cvpr/EderSLF20}         &                                                  & $45.6$ & $-$    & $-$    & $-$    & $-$         & $-$    & $-$    & $-$    & $-$    & $-$    & $-$    & $-$    & $-$    & $-$    \\
    \textbf{\textit{OURS}}   &                                                  & $52.9$ & $4.9$  & $63.9$ & $55.1$ & $83.1$ & $59.1$ & $40.2$ & $15.4$ & $57.7$ & $90.5$ & $33.8$ & $56.8$ & $70.9$ & $55.8$ \\
    \midrule
    HoHoNet~\cite{DBLP:conf/cvpr/0004SC21}         & \multirow{5}{*}{\rotatebox[origin=c]{90}{RGB-D}} & $56.7$ & $11.0$ & $63.7$ & $55.2$ & $88.9$ & $63.5$ & $45.2$ & $19.8$ & $67.5$ & $96.2$ & $37.4$ & $59.6$ & $74.3$ & $55.1$ \\
    PanoFormer~\cite{DBLP:conf/eccv/ShenLLNZZ22}      &                                                  & $57.0$ & $\boldsymbol{15.4}$ & $59.0$ & $54.9$ & $89.7$ & $66.1$ & $45.9$ & $20.1$ & $72.1$ & $97.2$ & $32.3$ & $62.5$ & $74.8$ & $51.5$ \\
    CBFC~\cite{DBLP:conf/wacv/ZhengLNLSZ23}            &                                                  & $56.7$ & $-$    & $-$    & $-$    & $-$         & $-$    & $-$    & $-$    & $-$    & $-$    & $-$    & $-$    & $-$    & $-$    \\
    Tangent~\cite{DBLP:conf/cvpr/EderSLF20}         &                                                  & $52.5$ & $-$    & $-$    & $-$    & $-$         & $-$    & $-$    & $-$    & $-$    & $-$    & $-$    & $-$    & $-$    & $-$    \\
    \textbf{\textit{OURS}}   &                                                  & $55.5$ & $7.9$  & $64.6$ & $56.1$ & $85.9$ & $69.3$ & $41.6$ & $17.5$ & $58.4$ & $96.0$ & $39.1$ & $61.4$ & $71.9$ & $51.6$ \\
    \midrule
    \textbf{\textit{OURS}}   &  \rotatebox[origin=c]{90}{ RGB-H }               & $\boldsymbol{60.6}$ & $10.8$ & $67.9$ & $59.0$ & $\boldsymbol{91.0}$ & $74.3$ & $53.1$ & $23.9$ & $68.1$ & $97.8$ & $43.3$ & $65.8$ & $76.0$ & $56.9$ \\
    \midrule
    \textbf{\textit{OURS}}   &  \rotatebox[origin=c]{90}{ RGB-N }               & $58.2$ & $10.8$ & $62.5$ & $57.6$ & $88.6$ & $71.0$ & $46.5$ & $20.2$ & $66.4$ & $97.4$ & $39.2$ & $64.1$ & $74.5$ & $58.4$ \\
    \midrule
    \textbf{\textit{OURS}}   &  \rotatebox[origin=c]{90}{ RGB-D-H }             & $60.0$ & $8.0$  & $67.3$ & $58.2$ & $90.6$ & $71.8$ & $49.5$ & $\boldsymbol{25.0}$ & $64.7$ & $97.8$ & $46.8$ & $65.9$ & $75.1$ & $59.4$ \\
    \midrule
    \textbf{\textit{OURS}}   &  \rotatebox[origin=c]{90}{ RGB-D-N }             & $59.4$ & $5.7$  & $\boldsymbol{77.6}$ & $\boldsymbol{65.7}$ & $90.4$ & $\boldsymbol{76.0}$ & $\boldsymbol{54.2}$ & $4.6$  & $\boldsymbol{81.9}$ & $97.9$ & $\boldsymbol{53.6}$ & $\boldsymbol{71.9}$ & $67.3$ & $\boldsymbol{69.0}$ \\
    \midrule
    \textbf{\textit{OURS}}   &  \rotatebox[origin=c]{90}{ RGB-N-H }             & $60.2$ & $7.8$  & $67.9$ & $59.3$ & $90.5$ & $73.2$ & $50.8$ & $22.8$ & $64.9$ & $\boldsymbol{98.1}$ & $44.5$ & $67.7$ & $\boldsymbol{76.3}$ & $59.3$ \\
    \bottomrule
  \end{tabular}
  \caption{Per-class results (\%) on the 3-fold validation of the Stanford2D3DS~\cite{DBLP:journals/corr/ArmeniSZS17} benchmark.}
  \label{tab:stanford2d3dsperclass}
\end{table*}

%%%%%%%%% REFERENCES
{\small
\bibliographystyle{ieee_fullname}
\bibliography{egbib.bib}
}

\end{document}